\documentclass[journal]{IEEEtran}
\ifCLASSINFOpdf
\else
\fi

\usepackage{graphicx}
\usepackage{times}
\usepackage{soul}
\usepackage{url}
\usepackage{subfigure}
\usepackage{caption}
\usepackage[hidelinks]{hyperref}
\usepackage[utf8]{inputenc}
\usepackage{amsmath,amssymb}
\usepackage{booktabs}
\usepackage{algorithm}
\usepackage{algorithmic}
\usepackage{mathtools}
\usepackage{array}
\usepackage{multirow}
\usepackage{stmaryrd}
\usepackage[misc]{ifsym}
\usepackage{booktabs}
\usepackage{tabularx}
\usepackage{xcolor}
\begin{document}
\title{Unsupervised Few-Shot Continual Learning for Remote Sensing Image Scene Classification}

\author{Muhammad Anwar~Ma'sum,
        Mahardhika~Pratama,~\IEEEmembership{Senior Member,~IEEE}
        , Ramasamy~Savitha,~\IEEEmembership{Senior Member,~IEEE}, Lin~Liu, Habibullah, Ryszard~Kowalczyk% <-this % stops a space
\thanks{M. A. Ma'sum, M. Pratama, L. Liu, Habibullah, R. Kowalcyzk are with STEM, University of South Australia, Adelaide, South Australia, Australia. R. Savitha is with I2R, A*Star, Singapore. R. Kowalczyk is also with System Research Institute, Polish Academy of Sciences.}% <-this % stops a space
\thanks{M. A. Ma'sum and M. Pratama share equal contributions}% <-this % stops a space
\thanks{Manuscript received April 19, 2005; revised August 26, 2015.}}


\markboth{Journal of \LaTeX\ Class Files,~Vol.~14, No.~8, August~2015}%
{Shell \MakeLowercase{\textit{et al.}}: Bare Demo of IEEEtran.cls for IEEE Journals}



\maketitle

\begin{abstract}
A continual learning (CL) model is desired for remote sensing image analysis because of varying camera parameters, spectral ranges, resolutions, etc. There exist some recent initiatives to develop CL techniques in this domain but they still depend on massive labelled samples which do not fully fit remote sensing applications because ground truths are often obtained via field-based surveys. This paper addresses this problem with a proposal of unsupervised flat-wide learning approach (UNISA) for unsupervised few-shot continual learning approaches of remote sensing image scene classifications which do not depend on any labelled samples for its model updates. UNISA is developed from the idea of prototype scattering and positive sampling for learning representations while the catastrophic forgetting problem is tackled with the flat-wide learning approach combined with a ball generator to address the data scarcity problem. Our numerical study with remote sensing image scene datasets and a hyperspectral dataset confirms the advantages of our solution. \textcolor{orange}{Source codes of UNISA are shared publicly in \url{https://github.com/anwarmaxsum/UNISA} to allow convenient future studies and reproductions of our numerical results.}
\end{abstract}


% \fi
\IEEEpeerreviewmaketitle

\section{Introduction}
\IEEEPARstart{T}{he} use of deep learning (DL) has been widespread in the domain of remote sensing (RS) such as earth observation, disaster monitoring, urban planning, etc. because it adopts an end-to-end training process removing the application of hand-crafted features. This progress is mainly driven by the operations of advanced satellite technologies facilitating remote sensing imaging technologies for convenient data collections. Nonetheless, RS domains are known to be rapidly changing because of temporal, spatial and spectral resolution limitations , thus calling for continual learning (CL) solutions, e.g., while earth observation sensors obtain images continuously over time, the land cover/ land use classes are made available sequentially \cite{Bhat2021CILEANETCI}. RS applications characterize four unique problems \cite{Ye2022BetterMB}: \textit{data openness}, \textit{temporal openness}, \textit{spatial openness} and \textit{spectral openness}. Naive retraining approaches impose considerable complexities which do not scale well for continuous environments. 

\begin{figure*}[h!]
\centering
\includegraphics[width=0.95\textwidth]{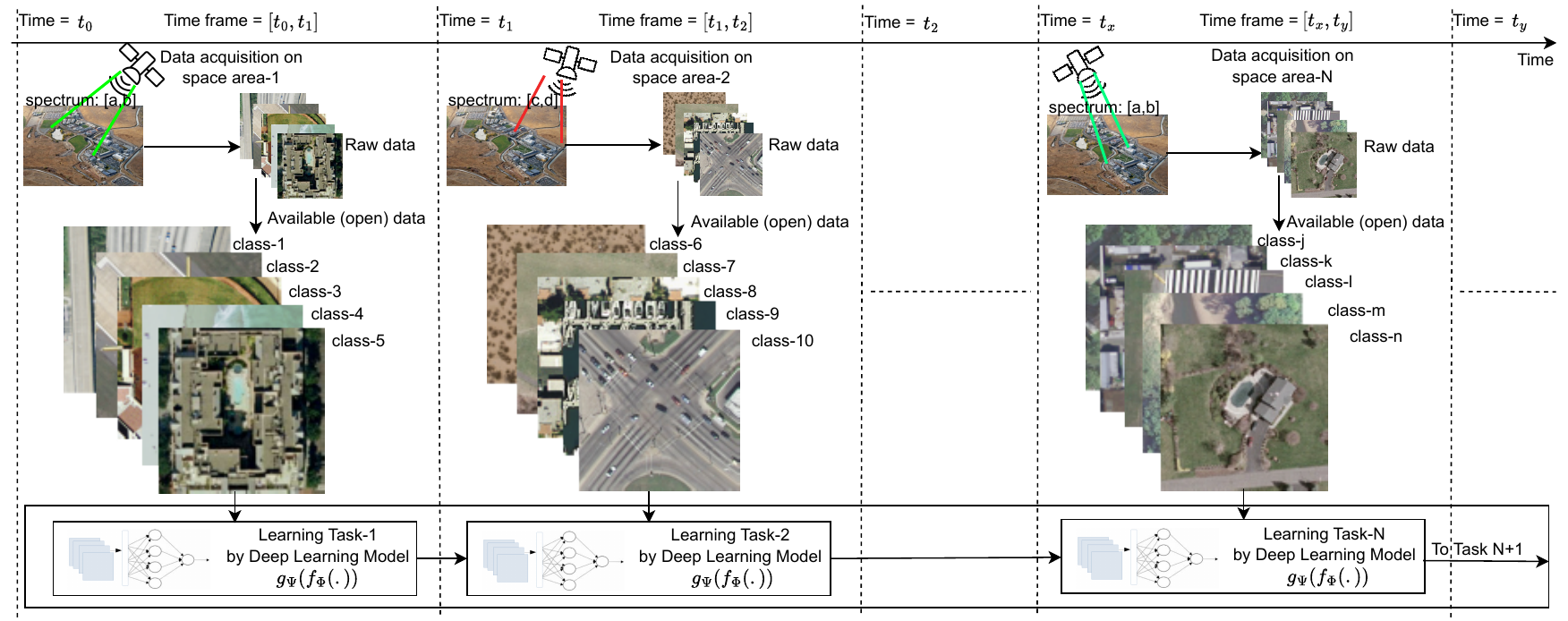}
\caption{\textcolor{red}{Visualization of Continual Learning for Remote Sensing Scene Classification}}
\label{Fig:cl_for_remote_sensing}
\end{figure*}

Continual learning (CL) is a research area of growing interests in the RS domain where several works have been recently proposed in the literature \cite{Zhai2019LifelongLF,Ammour2022ContinualLU,Lu2021LILLI,Ye2022BetterMB,Bhat2021CILEANETCI}. The concept of model agnostic meta learning (MAML) \cite{Finn2017ModelAgnosticMF} is implemented in \cite{Zhai2019LifelongLF} for CL of RS images. Lightweight incremental learning (LIL) approach is proposed in \cite{Lu2021LILLI} for continual RS image classification. \cite{Ammour2022ContinualLU} presents the notion of two trainable sub-networks and \cite{Bhat2021CILEANETCI} is based on the curriculum learning. The concept of preserving network and transient network is developed in \cite{Ye2022BetterMB} for CL of RS image scene classifications. Nonetheless, these works require a lot of labelled samples which do not fit to the RS applications. Labelling cost is often expensive in the RS domains because ground truth is based on the field-based survey. Fig. \ref{Fig:cl_for_remote_sensing} visualizes the case of CL for RS images. The number of classes cannot be fixed at once due to temporal, spatial and spectral resolution limitations leading to the case of CL, which should learn a sequence of new target classes without the catastrophic forgetting problem.

Some attempts have been made to address the problem of labelling costs in the RS area. \cite{Zhang2021TopologicalSA} proposes the concept of CNN and GCN for cross-scene hyper-spectral image classifications. \cite{Zhang2022SingleSourceDE} offers a single-source domain generalization framework in realm of cross-scene hyper-spectral image classifications based on the supervised contrastive learning framework. Another domain generalization framework is proposed in \cite{Zhang2022LanguageAwareDG} integrating linguistic modal information. \cite{Zhang2022HyperspectralAL} combines the hyper-spectral data and the Lidar data for collaborative land-cover classifications. Albeit these recent advances in handling the low label situation, these works do not yet address the CL problem where a model has to cope with a sequence of different learning tasks. In addition, the problem of limited data remains an uncharted territory.

How to handle continual learning problems with limited samples is known as few-shot continual learning (FSCL) \cite{Tao2020FewShotCL} and has attracted significant research interests in realm of CL. \cite{Tao2020FewShotCL} proposes the idea of neural gas, while \cite{Mazumder2021FewShotLL} puts forward the concept of session trainable parameters. \cite{Zhang2021FewShotIL} integrates the graph attention network to boost the performance of base task, while \cite{Chen2021IncrementalFL} addresses the FSCL problem using the concept of incremental vector quantization. \cite{Shi2021OvercomingCF} offers the idea of flat learning concept and \cite{Kalla2023S3CSS} proposes the idea of stochastic classifier. All of these works have not been validated with RS data. In addition, they still assume the presence of fully labelled training data albeit the low sample constraint. Note that \cite{Zhang2022GraphIA} does not address the CL problem although it deals with cross-domain few-shot learning problems. \textcolor{blue}{The motivations of this paper are outlined as follows:
\begin{itemize}
    \item Existing works in the CL for RS data do not yet address the problem of data scarcity. That is, they suffer from the over-fitting problem in the case of few-shot continual learning. 
    \item Although the problem of FSCL has been studied in some recent works, these approaches are not yet targeted for the RS data possessing unique characteristics. 
    \item Existing FSCL works assume the presence of fully labelled samples, which may not hold in many practical environments. In practise, labels are scarce and expensive to obtain. 
\end{itemize}}

This paper proposes unsupervised flat-wide learning approach (UNISA) for unsupervised few-shot continual learning (UFSCL) of RS image scene classification. UFSCL can be perceived as an extension of few-shot continual learning (FSCL) \cite{Tao2020FewShotCL} where no labelled samples are used for model updates. Very few samples are only made available for class inferences. UNISA is developed from the concept of deep clustering network where the representation training process is driven with the concept of prototype scattering and positive sampling \cite{Huang2021LearningRF} avoiding the class collision issue while attaining uniform representation. This strategy combines the concept of contrastive learning and non-contrastive learning where the prototype scattering loss assures the uniformity of representation while the positive sampling loss overcomes the class collision issue. The catastrophic forgetting problem is resolved with the union of flat learning concept \cite{Shi2021OvercomingCF} and wide learning concept \cite{Cha2021CPRCR} seeking and maintaining the flat-wide local minima regions. That is, the base learning task explores the flat-wide minima regions via noise perturbation while clamping network parameters to stay aroud the flat-wide local optimum regions during the few-shot learning tasks. The idea of ball generator \cite{Sun2021MEDAMW} is implemented during the few-shot learning tasks as a feature-level data augmentation approach to address the problem of data scarcity. 

\textcolor{blue}{This paper contains four major contributions:
\begin{enumerate}
    \item we propose a challenging but practical problem, unsupervised few-shot continual learning (UFSCL) problems. This problem emphasizes the low label problem in the few-shot continual learning (FSCL) setting where no labelled samples are offered for model updates;
    \item unsupervised flat-wide learning (UNISA) approach is proposed as a solution of the UFSCL problem. To the best of our knowledge, existing FSCL approaches exclude the low-label problem and thus are not directly applicable to the UFSCL setting;
    \item we propose a novel flat-wide learning approach based on the concept of prototype scattering and positive sampling. This approach goes one step ahead of current flat-wide learning approach \cite{Masum2023FewShotCL} only designed in the fully supervised learning contexts;
    \item a new joint loss functions for UFSCL problems is put forward. The representation learning component aims to extract meaningful representation without any labels and the ball data generator component is meant to enrich latent space representation while the wide learning component and the regularization term protect against the catastrophic forgetting problem. 
\end{enumerate}} 
Rigorous numerical study has been executed in which it includes comparisons of prior arts. It is shown that UFSCL is a highly challenging problem for existing solutions of few-shot continual learning problems. UNISA outperforms prior arts with substantial gaps.
%2) unsupervised flat-wide learning (UNISA) approach is proposed as a solution of the UFSCL problem; 3) new joint loss functions integrating the flat learning concept and the wide learning concept under the framework of prototype scattering and positive sampling are proposed; 4) source codes of UNISA are shared publicly in \url{https://github.com/anwarmaxsum/UNISA} to allow convenient future studies and reproductions of our numerical results. Rigorous numerical study has been executed in which it includes comparisons of prior arts. It is shown that UFSCL is a highly challenging problem for existing solutions of few-shot continual learning problems. UNISA outperforms prior arts with substantial gaps.

\section{Related Works}
\subsection{Continual Learning}
Continual Learning (CL) has rapidly progressed in the literature where the major challenge is to overcome the catastrophic forgetting (CF) problem when learning new tasks, thus making possible for knowledge accumulation overtime. Existing approaches consist of three groups \cite{Parisi2019ContinualLL}: regularization-based approach, architecture-based approach and memory-based approach. Regularization-based approach \cite{Li2016LearningWF,Kirkpatrick2017OvercomingCF,Zenke2017ContinualLT,Aljundi2018MemoryAS,Schwarz2018ProgressC,Paik2020OvercomingCF,Mao2021ContinualLV,Cha2021CPRCR} relies on a penalty term constraining important parameters from significant deviations to relieve the CF problem. This approach, however, does not scale well for long sequences of tasks and depends on the task ID. The architecture-based approach is based on a growing network structure followed by isolation of old parameters to combat the CF problem \cite{Rusu2016ProgressiveNN,Yoon2018LifelongLW,Li2019LearnTG,Xu2021AdaptivePC,Ashfahani2022UnsupervisedCL,Pratama2021UnsupervisedCL,Rakaraddi2022ReinforcedCL}. This approach imposes costly computational burdens while still relying on the task IDs. The memory-based approach is based on an episodic memory storing old data samples to be interleaved with new samples when learning new tasks \cite{Rebuffi2017iCaRLIC,LopezPaz2017GradientEM,Ebrahimi2020AdversarialCL,Pham2021ContextualTN,VinciusdeCarvalho2022ClassIncrementalLV,Buzzega2020DarkEF,Dam2022ScalableAO,Masum2023AssessorGuidedLF}. Although these approaches offer strong performances with the absences of task IDs, they are not feasible for the FSCL problem due to the few sample limitation. The application of episodic memory in such setting likely leads all samples of a task to be stored in the memory, thus being against the CL spirit. CL has also attracted research attentions in the RS field as shown in \cite{Zhai2019LifelongLF,Ammour2022ContinualLU,Lu2021LILLI,Ye2022BetterMB,Bhat2021CILEANETCI}. These approaches require a lot of labelled data samples and on the other hand RS data usually do not carry long historical information. \textcolor{blue}{Recent works on the CL problem utilize a frozen pre-trained ViT backbone and learned additional parameters called prompts to adapt to the sequence of tasks \cite{wang2022learning,wang2022dualprompt,smith2023coda}. The other studies enhance the prompt learning with a self-evolving structure to accommodate the increasing of learned knowledge \cite{kurniawan2024evolving}. Another approach enhanced it with language models as additional information\cite{khan2023introducing,roy2024convolutional}. Even though they are proven in the CL problem, the prompt-based approach is not yet validated in FSCL integrating a data scarcity challenge.} 

\subsection{Few-Shot Continual Learning}
Few-Shot Continual Learning (FSCL) \cite{Tao2020FewShotCL} extends CL where it imposes a small sample constraint. That is, there exist a base learning task where many samples are supplied and a sequence of few-shot learning tasks where very few samples are accessible under the N-way K-shot configuration, i.e., every task carries $N$ classes where each class only has $K$ samples. \cite{Tao2020FewShotCL} utilizes the concept of neural gas, \cite{Mazumder2021FewShotLL} puts forward the idea of session trainable parameters, \cite{Zhang2021FewShotIL} puts forward the idea of graph attention network, \cite{Chen2021IncrementalFL} utilizes the concept of incremental vector quantization and \cite{Shi2021OvercomingCF} applies the flat minima concept. Note that the concept of flat learning in \cite{Shi2021OvercomingCF} does not involve the concept of wide learning and performs poorly under small base tasks because of inaccurate estimations of flat regions. In addition, \cite{Shi2021OvercomingCF} compromises the plasticity of a model when learning new concepts. \cite{Kalla2023S3CSS} combines the ideas of self-supervised learning and stochastic classifier. The FSCL problem does not address the issue of labeling costs and assumes the presence of fully labelled samples. This issue does not fit well for RS applications where the labelling cost is expensive.  \textcolor{blue}{Recent works in FSCL try to calibrate the classes' prototypes and use them as classifiers to minimize forgetting \cite{ahmed2024orco, wang2024few,yang2023neural,Paeedeh2024FewshotCI}.  The other approaches adjust model weight e.g. with geometric transformation or adding sub-networks \cite{kim2022warping,yoon2023soft}. Note that the methods are trained in a supervised way, both in the base task and the few shot tasks, and require abundant labeled samples for base task training. Thus, they are not yet proven in the case of low label in the base task.}

\subsection{Unsupervised Continual Learning}
Unsupervised continual learning (UCL) \cite{Rao2019ContinualUR} addresses the issue of labelling cost in CL where the training process is performed with the absence of any labelled samples. Small fractions of labels can be made available only for class inferences. \cite{Smith2019UnsupervisedPL,Ashfahani2022UnsupervisedCL,Pratama2021UnsupervisedCL} cope with this problem via deep clustering networks with a self-evolving property. The key difference with our proposed UFSCL in this paper lies in the access of a lot of unlabelled samples for UCL whereas the UFSCL does not enjoy such access. 
\subsection{Continual Learning in RS}
Continual learning (CL) has triggered active researches in the RS domain. The concept of MAML \cite{Finn2017ModelAgnosticMF} is integrated in \cite{Zhai2019LifelongLF} for lifelong learning of RS images. The concept of lightweight incremental learning (LIL) is proposed in \cite{Lu2021LILLI} for CL of RS images. \cite{Ammour2022ContinualLU} adopts the idea of two trainable sub-networks while \cite{Bhat2021CILEANETCI} proposes the notion of curriculum learning. The concept of preserving and transient networks is proposed in \cite{Ye2022BetterMB}. \textcolor{green}{In \cite{Dang2023DistributionRA}, distribution reliability assessment based on incremental learning (DRAIL) is proposed for automatic target recognition. A novel exemplar selection is proposed in \cite{Fu2024ClassIncrementalRO} for class-incremental learning algorithm and applied to the RS data. Synthetic Aperture Radar Automatic Target Recognition (SARATR) is proposed in \cite{Wang2023FewShotCS} and based on the few-shot class-incremental learning algorithm.} All these works are not compatible for the UFSCL problem proposed here because they are still designed for ordinary CL problems excluding low label-and low data-situations in the CL problems.

\begin{figure*}[h!]
\centering
\includegraphics[width=0.95\textwidth]{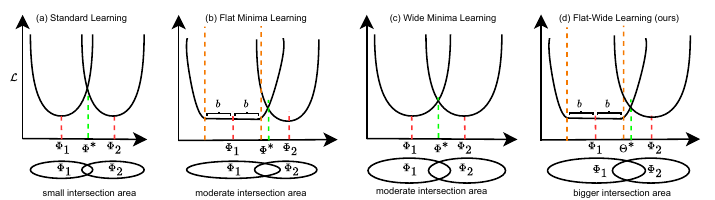}
\caption{\textcolor{blue}{Visualization of Model Parameters in Flat Wide Learning (d) compared to Standard Learning (a), Flat Minima Learning (b), and Wide Minima Learning (d)}}
\label{Fig:flat_wide}
\end{figure*}

\section{Preliminaries}
\subsection{Problem Definition}
\textcolor{orange}{Unsupervised few-shot continual learning (UFSCL) is defined as a sequential learning problem over a sequence of tasks, $\mathcal{T}_1,...,\mathcal{T}_K, k\in\{1,...,K\}$ under the presence of both label and data scarcities. $K$ stands for the number of tasks unknown in practise. Unlike conventional FSCL problem \cite{Tao2020FewShotCL}, a model is fed only by unlabelled samples $\mathcal{T}_k=\{x_i^{k}\}_{i=1}^{|T_k|}$ where $|.|$ stands for the cardinality symbol and $x_i^{k}\in X$ denotes the $i-th$ image of the $k-th$ task. A very small subset of labels can be accessed only for class inferences $Y_k=\{y_i\}_{i=1}^{|Y_k|},|Y_k|<<|\mathcal{T}_k|$ where $y_i\in\mathcal{Y}$ denotes a class label of the $i-th$ image and $Y_k$ stands for the label set of the $k-th$ task. All images are drawn from the same domain $\mathcal{X}\times\mathcal{Y}\in\mathcal{D}$. We only focus on the challenging class-incremental learning setting \cite{Ven2019ThreeSF} here where a model is configured under a single head structure with the absence of any task IDs. That is, each task carries different and non-overlapping label sets $\forall_{k,(k-1)}Y_k\cap Y_{k-1}\in\emptyset$.} 

As with the FSCL problem, the UFSCL problem comprises a base learning task $\mathcal{T}_1$ and streaming few-shot learning tasks $\mathcal{T}_k$. There exist large samples of the base learning task $\mathcal{T}_1$ whereas the data scarcity issue is seen in the few-shot learning tasks $\mathcal{T}_k$. That is, the few-shot task $\mathcal{T}_k$ is configured in the N-way K-shot setting, i.e., it carries N classes where each class contain only K samples. We explore the case of $K\in[1,5]$. 

UNISA is structured as a deep clustering network composed of a feature extractor $z=f_{\Phi}(x)\in\Re^{D}:x\rightarrow z$ mapping an input image $x\in\Re^{w\times h\times c}$ to a feature space, fully connected layers $\hat{z}=g_{\Psi}(z)\in\Re^{\hat{D}}: z\rightarrow \hat{z}$, and a clustering module $\hat{y}=\kappa_{C}(\hat{z}):\Re^{\hat{D}}\rightarrow \Delta_M$ delivering final predictions in which $M$ denotes the output dimension.

\begin{figure*}
\centering
\includegraphics[width=0.95\textwidth]{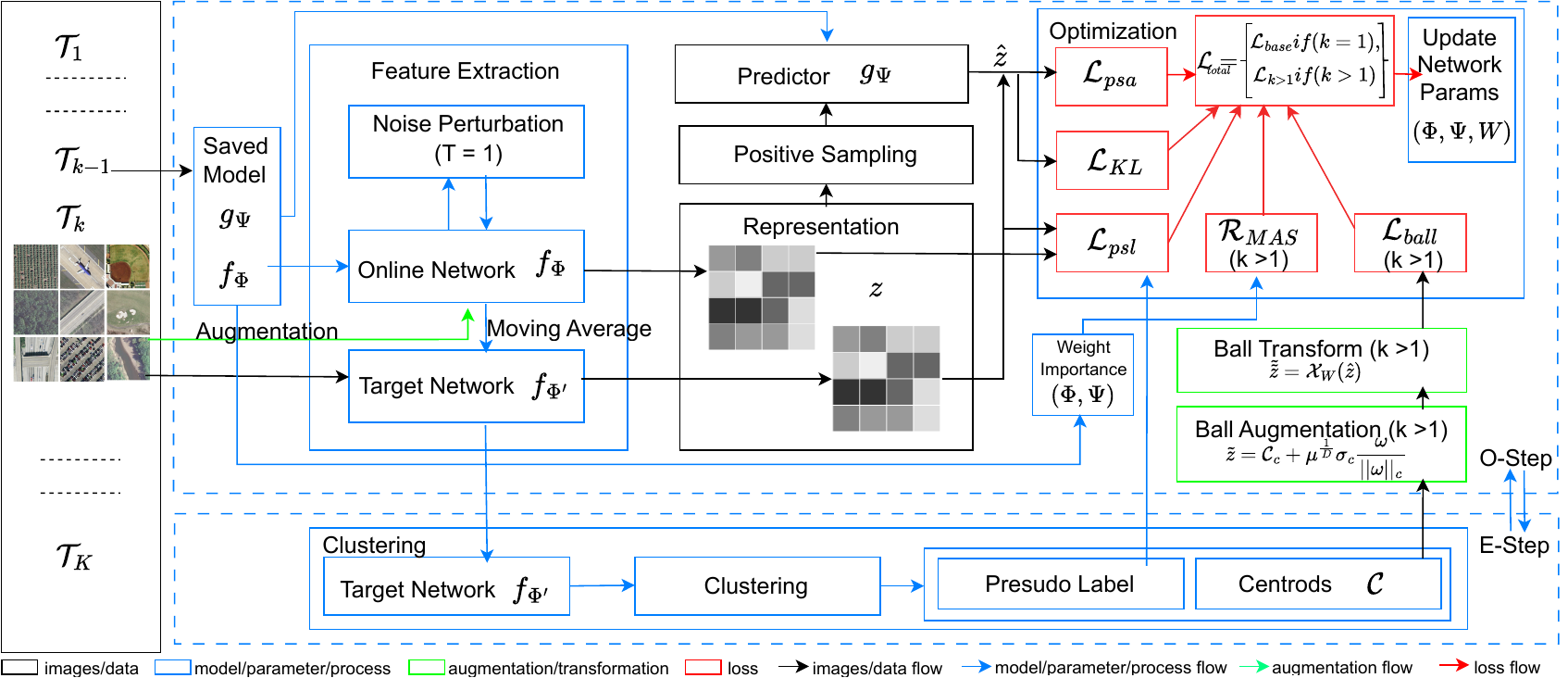}
\caption{\textcolor{blue}{Workflow of the Proposed Method (UNISA): UNISA is trained minimizing a joint loss function of the prototype scattering loss, the positive sampling alignment loss, the wide local minimum loss, the ball generator loss and the regularization term. The data augmentation procedure is implemented to address the data scarcity problem.}}
\label{Fig:UNISA_method}
\end{figure*}

\textcolor{green}{\subsection{Our Solution: Unsupervised Flat-Wide Learning Approach}}
\textcolor{green}{In this section, we introduce the main concept of our proposed method i.e. unsupervised flat-wide learning illustrated in figure \ref{Fig:flat_wide}. Given a deep learning model $g_{\Psi}(f_{\Phi}(.))$ where $f$ and $g$ is feature extractor and classifier that parameterized by $\Phi$ and $\Psi$ respectively. Standard FSCL learning optimizes $\Phi$ to recognize the learned classes from the base task ($T_1$) until the current task ($T_2$). However, since the tasks have different feature distributions, the optimal intersection of optimal parameters for $T_1$ and $T_2$ is small as shown in figure \ref{Fig:flat_wide} (a). Flat minima learning shown in figure \ref{Fig:flat_wide} (b) tries to search the flat region on the base task by perturbing $\Phi_1$ with noise in the range [-b,b] then forced to clamp the optimal parameter $\Phi^*$ into the flat region. Wide minima learning strives to enlarge the insertion between $\Phi_1$ and $\Phi_2$ by scratching both loss curves as shown in figure \ref{Fig:flat_wide} (c). Our proposed method i.e. Flat-Wide Learning leverages both properties. In the base task, It finds a flat region while simultaneously scratching its loss curve. In the few-shot task, the model keep applying the wide minima principle, so the intersection between $\Phi_1$ and $\Phi_2$ is bigger than the previous three approaches. The higher intersection area implies the model has a higher chance of recognizing classes in both tasks.}

\textcolor{green}{The second uniqueness is that our proposed method is trained in an unsupervised way, while the previous three approaches are trained in a supervised way. Instead of the actual label, our approach utilizes a pseudo-label produced by an unsupervised learning mechanism i.e. clustering to optimize $\Phi$. Third, our proposed method creates multiple representations for each input. In each iteration, given an input $x$, our method creates representations $f_{\Phi}(x)$ and $f_{\Phi'}(x)$ where $\Phi'$ is an alternate version of $\Phi$ by applying a moving average operation. This enrichment of representation plays an important role in optimizing $\Phi$. These last two principles try to answer the data scarcity challenge where only a small number of labeled data is available. The details of our proposed method are presented in the following section.}

\section{Base Learning Session $k=1$}
This section outlines the procedure of base learning session of UNISA whose goal is to solicit flat-wide local minima to be retained throughout its learning process via representation learning of the prototype scattering and positive sampling losses.
\subsection{Representation Learning}
The clustering-friendly latent space of UNISA is attained via a representation learning method minimizing the prototype scattering and positive sampling losses \cite{Huang2021LearningRF}. This concept combines the concept of contrastive and non-contrastive learning under one roof taking advantages of their strengths while eliminating their bottlenecks. Notwithstanding that the contrastive learning strategy \cite{Wu2018UnsupervisedFL} using the InfoNCE loss allows uniform representations, it suffers from the class collision issue where negative samples of the same clusters are wrongly pushed away. On the other hand, such drawback is absent in the non-contrastive learning technique but it often results in trivial solutions leading to too many empty clusters. Let $x$ be an instance where $x^{+}$ is its augmented version via random augmentation and $\{x_1^{-},...,x_{m}^{-}\}$ be a set of $m$ negative instances, the InfoNCE loss can be written as follows: 
\begin{equation}
\begin{split}
        \mathcal{L}_{INCE}\approx\underbrace{-\frac{g_{\Psi}(f_{\Phi}(x)^{T})g_{\Psi}(f_{\Phi}(x^{+}))}{\tau}}_{alignment}\\+\underbrace{\log\sum_{i=1}^{m}\exp{(\frac{g_{\Psi}(f_{\Phi}(x)^{T})g_{\Psi}(f_{\Phi}(x^{-}))}{\tau})}}_{uniformity}
\end{split}
\end{equation}
pulling together the positive pair $(x,x^{+})$, the alignment term, while pushing away $x$ from a set of $m$ negative instances, the uniformity term. The absence of cluster representation causes the class collision issue due to the uniformity term. The uniformity term is discarded in the non-constrastive learning technique, thus avoiding the class collision issue but suffers form non-uniform representations. 

The prototype scattering loss works directly on the cluster level using the contrastive strategy. Suppose that there exists $N_C$ clusters, one cluster is assigned as a positive cluster while the remainder $N_C-1$ clusters are selected as negative prototypes. $N_C$ clusters are constructed at the bottleneck layer of the fully connected module $\{C_1,C_2,....,C_{N_C}\}$ and at the same time their augmented versions are also created using the random augmentation approach $\{C_{1}^{'},C_{2}^{'},...,C_{N_C}^{'}\}$. The prototype scattering loss function is formalized as follows: 
\begin{equation}\label{psl}
    \mathcal{L}_{psl}=\frac{1}{N_C}\sum_{c=1}^{N_C}-\frac{C_{c}^{T}C_c^{'}}{\tau}+\frac{1}{N_C}\sum_{s=1}^{N_C}\log\sum_{\underset{c\neq s}{c=1}}^{N_C}\exp{(\frac{C_{s}^{T}C_{c}}{\tau})}
\end{equation}
\eqref{psl} guarantees uniform representations and well-separated clusters without any risk of the class collision issue because the negative instance is of different clusters. 

The positive sampling alignment approach is developed to improve within-cluster compactness while further overcoming the class collision problem. This is achieved by aligning the neighbouring examples with another view of a sample. Note that neighbouring samples are truly positive and belong to the same class \cite{Huang2021LearningRF}. Neighbouring samples themselves are constructed with the help of the Gaussian distribution controlling the influence zone of a sample to be treated as positive samples. The positive alignment loss is written as follows:
\begin{equation}\label{psa}
\begin{split}
\mathcal{L}_{psa}=||g_{\Psi}(f_{\Phi}(x+\epsilon\sigma))-f^{'}(x^{+})||_2^{2}
\end{split}
\end{equation}
where $\epsilon\backsim\mathcal{N}(0,1)$ and $\sigma$ is a predefined constant. $f^{'}(.)$ is a target network, momentum updated. \eqref{psa} should not cause any class collision issue because neighboring samples of an instance should be positive samples.  

This strategy is implemented in the EM framework where the k-means clustering strategy is applied in the E step followed by the fine-tuning mechanism during the M step minimizing the proposed loss function. The target network is updated using moving average formula as \cite{Huang2021LearningRF,Chen2020ExploringSS} $\phi^{'}=m\phi^{'}+(1-m)\phi$ and used in the inference phase. 

\subsection{Flat-Wide Base Learning Phase $k=1$}
The base learning phase is devised to capture flat-wide local optimum regions in which the losses are still minimized while coping with the CF problems. The flat learning concept itself \cite{Shi2021OvercomingCF} does not suffice because it risks on too narrow local optimum regions which do not scale well for large-scale problems. As a result, the flat learning concept usually works well with large base tasks and need to be supplemented with an episodic memory to address the CF problem. The concept of wide learning \cite{Cha2021CPRCR} is integrated here to drive the flat regions to be wide, thereby increasing the chance of overlapping regions across all tasks to be discovered. The flat local optimum regions $\Phi-b\leq\Phi\leq\Phi+b$ can be found by introducing small random noises $-b\leq\epsilon_t\leq+b$ to model's parameters multiple times such that similar but different losses are returned. Such losses can be minimized simultaneously to discover the flat minimum regions. Formally speaking, our interest is to minimize the expected loss function w.r.t the joint distribution of data $z$ and noise $\epsilon,-b\leq\epsilon\leq+b$:
\begin{equation}
\begin{split}
    R(\Phi)=\int_{\Re^{d_{\epsilon}}}\int_{\Re^{d_{z}}}\mathcal{L}(z;\Phi+\epsilon,\Psi)dP(z)dP(\epsilon)\\=\mathbb{E}[\mathcal{L}(z;\Phi+\epsilon,\Psi)]
\end{split}
\end{equation}
where $P(z),P(\epsilon)$ respectively stand for the data distribution and the noise distribution. In practice, expected losses are impossible to be minimized, so its empirical version is formulated as follows:
\begin{equation}\label{Lbase}
    \begin{split}
        \mathcal{L}_{k=1}=\frac{1}{T}\sum_{t=1}^{T}\mathcal{L}_{base}(z;\Phi+\epsilon_{t},\Psi)\\\mathcal{L}_{base}=\mathbb{E}_{(x)\backsim\mathcal{T}_{1}}[\mathcal{L}_{psl}(z;\Phi+\epsilon_t,\Psi,C)+\lambda_1\mathcal{L}_{psa}(z;\Phi+\epsilon_t,\Psi)\\+\lambda_2\mathcal{L}_{KL}(g_{\Psi}(f_{\Phi}(x)),P_{U})]+\lambda_{3}||C_{c}-C_{c}^{*}||_2
    \end{split}
\end{equation}
where $\lambda_1,\lambda_2,\lambda_3$ denote predefined constants. $\mathcal{L}_{KL}(.)$ is appended to promote wide local optima where latent features are projected to $P_{U}$, the uniform distribution. That is, the uniform distribution is selected because it offers the upper bound of divergence $\log{M}$. $C_{c},C_{c}^{*}$ denote the latent clusters before and after noise injections where the last term is included to avoid the cluster's drifts. Note that we follow \cite{Chen2020ExploringSS} where the target network $f_{\Phi^{'}}(.)$ is momentum updated to allow stable representations. That is, the online network $f_{\Phi}(.)$ is updated first minimizing \eqref{Lbase} and applied to the moving average formula to adjust the target network.

\section{Few-Shot Learning Phase $k>1$}
The few-shot learning phase features data scarcity issues where it is configured under the $N$-way $K$-shot configuration. UNISA applies the feature-space data augmentation strategy to overcome this issue while putting forward the regularization-based approach to handle the CF problem. 
\subsection{Ball Data Generator}
The concept of ball data generator \cite{Sun2021MEDAMW} is modified to fit images rather than texts and also in the unsupervised manner. Synthetic samples of the ball data generator are induced within the smallest enclosing balls of the feature space to prevent out of distributions. Since three exist very few labels, the ball $\zeta(C_c,\sigma_c)$ is constructed via clusters in the unsupervised manner and the episodic training phase as practised in the few-shot learning domain is absent in the FSCL problem. $(C_c,\sigma_c)$ denote the centroid and standard deviation of the $c-th$ cluster estimated via the $k$-means clustering procedure. A synthetic sample $\tilde{z}\in\Re^{\hat{D}}$ is generated as follows:
\begin{equation}
    \tilde{z}=C_c+u^{\frac{1}{D}}\sigma_c\frac{\omega}{||\omega||_2}
\end{equation}
where $u\backsim\mathcal{U}(0,1)$ denotes a shifting constant and $\omega\in\Re^{\hat{D}}\backsim\mathcal{N}(0,1)$ labels a random vector. $\tilde{z}$ is not directly injected to the training process rather fed to a projection module $\tilde{\tilde{z}}=\chi_{W}(\tilde{z})\in\Re^{\hat{D}}:\hat{z}\rightarrow \tilde{\tilde{z}}$ to generate stable features. The projection module is trained in such a way to avoid synthetic sample biases via the triplet loss. 
\begin{equation}
    \mathcal{L}_{ball}=\sum_{\underset{\tilde{\tilde{z}}\in C_i}{\tilde{\tilde{z}}\notin C_j}}\lambda_{4}\max\{0,d(\tilde{\tilde{z}},C_i)+r-d(\tilde{\tilde{z}},C_j)\}
\end{equation}
where $r,\lambda_4$ are predefined constants respectively. The goal is such that synthetic samples stay close to its centroid while pushing them away from other centroids where the distance margin is controlled by $r$. $C_i,C_j$ are the centroid of the $i-th$ and $j-th$ clusters respectively.  
\subsection{Regularization-based Approach}
UNISA relies on the regularization-based approach to combat the CF problem due to sequences of the few-shot learning task. We apply the memory aware synapses (MAS) \cite{Aljundi2018MemoryAS} which can be computed in the online and unsupervised manner. Suppose that $\Theta=\{\Phi,\Psi\}$, the MAS technique is derived as follows:
\begin{equation}\label{reg}
    \mathcal{R}_{MAS}= \lambda_{5}\sum_{i}\Gamma_{i}^{k-1}(\Theta_{i}^{k}-\Theta^{k-1})^{2}
\end{equation}
where $\Gamma$ is a parameter importance matrix and $\lambda_5$ is a predefined constant. MAS is preferred over EWC \cite{Kirkpatrick2017OvercomingCF} and SI \cite{Zenke2017ContinualLT}, because it is faster to compute than them and applicable for unsupervised scenarios. We do not impose any regularization to the transformation module $\chi_{W}(.)$ because it only focuses on the current concept. \eqref{reg} can be perceived as the $L_2$ regularization except the presence of the difference across two consecutive weight matrices $\Theta^{k-1},\Theta^{k}$. 
\subsection{Flat-Wide Few-Shot Learning Phases $k>1$}
The loss function of the few-shot learning phases is written as follows:
\begin{equation}
 \begin{split}   \mathcal{L}_{k>1}=\mathbb{E}_{(x)\backsim\mathcal{T}_{k}}[\mathcal{L}_{psl}(z;\Phi,\Psi,C)+\lambda_1\mathcal{L}_{psa}(z;\Phi,\Psi)+\\\lambda_2\mathcal{L}_{KL}(g_{\Psi}(f_{\Phi}(x)),P_U)+\lambda_{4}\mathcal{L}_{ball}]+\mathcal{R}_{MAS}
 \end{split}
\end{equation}
That is, the prototype scattering loss $\mathcal{L}_{psl}$ and the positive sampling alignment loss $\mathcal{L}_{psa}$ loss learn the current concept with the support of synthetic samples induced by the ball generator. The ball generator loss $L_{ball}$ is designed to produce quality samples while $L_{KL}$ exists to induce wide local optimum regions and drives flat regions to be wide. Last but not least, the regularization $\mathcal{R}_{MAS}$ addresses the CF problem. Note that the projection module is trained in the end-to-end manner alongside the online network. The flat region is retained by applying the clamping mechanism. That is, the network parameters are clamped when going beyond the flat regions. In addition, the target network is adjusted by the momentum strategy of that the online network. The workflow of UNISA is pictorially depicted in Fig. \ref{Fig:UNISA_method}. Pseudo-code of UNISA is provided in the supplemental document. 

% \begin{figure*}
% \includegraphics[width=1.0\textwidth]{UCMerceddataset.png}
% \caption{Visualization of UCMerced dataset}
% \label{Fig:ucmerceddataset}
% \end{figure*}
\begin{figure}
\includegraphics[width=0.49\textwidth]{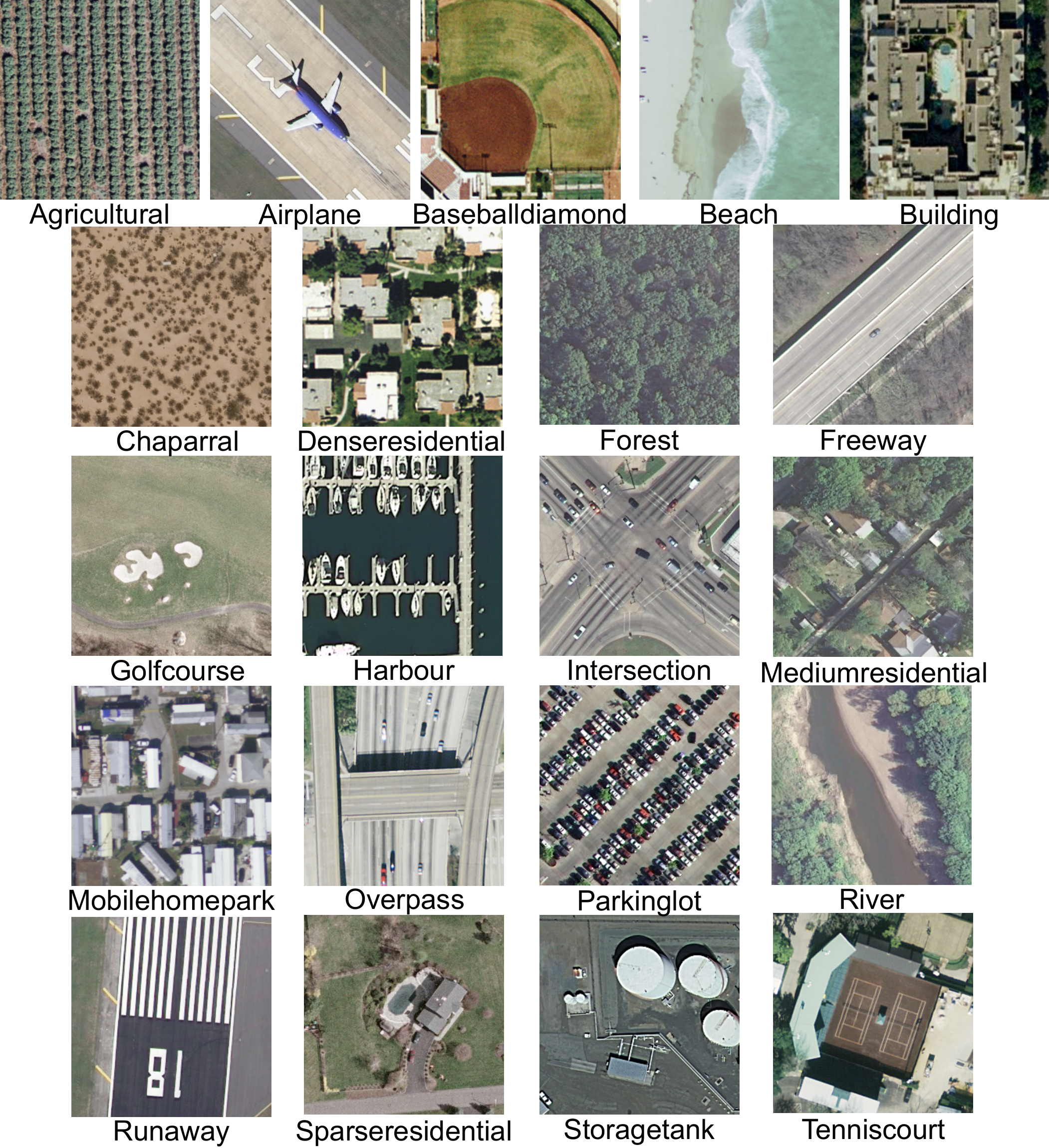}
\caption{\textcolor{blue}{Visualization of UCMerced dataset}}
\label{Fig:ucmerceddataset}
\end{figure}

\begin{table*}[]
\setlength{\tabcolsep}{0.21em}
\scriptsize
\centering
\begin{tabular}{l|c|cccccc|cccccc}
\hline
\multirow{3}{*}{Method} & \multirow{3}{*}{\#S} & \multicolumn{6}{c}{Split UCMerced}                                                                                                                     & \multicolumn{6}{c}{Split AID}                                                                                                                          \\ \hline
                        &                      & \multicolumn{4}{c}{Session}                           & \multirow{2}{*}{Avg} & \multirow{2}{*}{\begin{tabular}[c]{@{}c@{}}Gap to\\ UNISA\end{tabular}} & \multicolumn{4}{c}{Session}                           & \multirow{2}{*}{Avg} & \multirow{2}{*}{\begin{tabular}[c]{@{}c@{}}Gap to\\ UNISA\end{tabular}} \\ \cline{3-6} \cline{9-12}
                        &                      & 1           & 2           & 3           & 4           &                      &                                                                         & 1           & 2           & 3           & 4           &                      &                                                                         \\ \hline
Baseline\cite{Shi2021OvercomingCF}                & \multirow{9}{*}{5}   & 18.33$\pm$0.00          & 17.20$\pm$0.90          & 14.61$\pm$1.43          & 11.90$\pm$0.81          & 15.51$\pm$0.94          & -39.79$\pm$1.47                                                            & 18.06$\pm$0.00          & 12.87$\pm$0.69          & 10.01$\pm$0.29          & 7.90$\pm$0.43           & 12.21$\pm$0.43          & -49.21$\pm$0.71                                                            \\
F2M\cite{Shi2021OvercomingCF}                     &                      & 25.00$\pm$0.00          & 16.80$\pm$0.57          & 13.83$\pm$0.85          & 12.14$\pm$1.06          & 16.94$\pm$0.74          & -38.36$\pm$1.35                                                            & 16.49$\pm$0.00          & 13.38$\pm$0.25          & 10.10$\pm$0.17          & 9.67$\pm$0.37           & 12.41$\pm$0.24          & -49.01$\pm$0.61                                                            \\
FSLL\cite{Mazumder2021FewShotLL}                    &                      & 22.50$\pm$0.00          & 10.80$\pm$1.40          & 9.67$\pm$1.14           & 7.76$\pm$0.45           & 12.68$\pm$0.93          & -42.62$\pm$1.46                                                            & 25.78$\pm$0.00          & 12.13$\pm$2.02          & 8.34$\pm$0.71           & 7.15$\pm$0.39           & 13.35$\pm$1.09          & -48.07$\pm$1.23                                                            \\
ICARL\cite{Rebuffi2017iCaRLIC}                   &                      & 23.50$\pm$1.47          & 18.80$\pm$1.49          & 16.72$\pm$0.90          & 14.48$\pm$1.32          & 18.38$\pm$1.32          & -36.92$\pm$1.74                                                            & 16.41$\pm$0.31          & 14.58$\pm$0.73          & 12.47$\pm$0.64          & 10.15$\pm$0.71          & 13.40$\pm$0.62          & -48.02$\pm$0.84                                                            \\
Rebalance\cite{Hou2019LearningAU}               &                      & 12.50$\pm$0.23          & 10.07$\pm$0.22          & 8.22$\pm$0.12           & 7.24$\pm$0.21           & 9.51$\pm$0.20           & -45.79$\pm$1.15                                                            & 7.52$\pm$0.00           & 5.41$\pm$0.00           & 4.34$\pm$0.00           & 3.60$\pm$0.00           & 5.22$\pm$0.00           & -56.20$\pm$0.56                                                             \\
S3C\cite{Kalla2023S3CSS}                     &                      & 50.00$\pm$0.00          & 42.07$\pm$2.10          & 34.83$\pm$1.34          & 30.24$\pm$1.29          & 39.29$\pm$1.40          & -16.01$\pm$1.80                                                             & 49.27$\pm$0.00          & 34.08$\pm$1.09          & 27.94$\pm$1.03          & 23.92$\pm$0.47          & 33.80$\pm$0.79          & -27.62$\pm$0.97                                                            \\
SCN\cite{Ye2022BetterMB}                     &                      & 23.75$\pm$0.00          & 10.47$\pm$2.01          & 8.83$\pm$1.56           & 7.76$\pm$1.02           & 12.70$\pm$1.37          & -42.6$\pm$1.78                                                             & 36.33$\pm$0.00          & 13.50$\pm$1.28          & 11.77$\pm$1.27          & 12.15$\pm$0.62          & 18.44$\pm$0.95          & -42.98$\pm$1.10                                                            \\
\textcolor{orange}{ORCO\cite{ahmed2024orco}}                    &                      & 42.50$\pm$0.00          & 13.07$\pm$2.95          & 15.89$\pm$2.05          & 15.67$\pm$2.75          & 21.78$\pm$2.26          & -33.52$\pm$2.53                                                            & 31.94$\pm$0.00          & 14.85$\pm$1.88          & 10.61$\pm$1.80          & 12.73$\pm$1.58          & 17.53$\pm$1.52          & -43.89$\pm$1.62                                                            \\
\textbf{UNISA(Ours)}    &                      & \textbf{67.50$\pm$0.00} & \textbf{58.27$\pm$1.51} & \textbf{51.67$\pm$0.77} & \textbf{43.76$\pm$1.50} & \textbf{55.3$\pm$1.13}  & \textbf{0.00$\pm$0.00}                                                     & \textbf{85.91$\pm$0.00} & \textbf{63.83$\pm$0.76} & \textbf{52.23$\pm$0.52} & \textbf{43.71$\pm$0.63} & \textbf{61.42$\pm$0.56} & \textbf{0.00$\pm$0.00}                                                     \\  \hline
Baseline\cite{Shi2021OvercomingCF}                & \multirow{9}{*}{4}   & 18.33$\pm$0.00          & 17.33$\pm$1.12          & 10.44$\pm$2.83          & 8.86$\pm$2.22           & 13.74$\pm$1.88          & -38.37$\pm$2.62                                                            & 18.06$\pm$0.00          & 13.00$\pm$0.84          & 9.49$\pm$0.27           & 8.20$\pm$0.09           & 12.19$\pm$0.44          & -49.26$\pm$0.70                                                            \\
F2M\cite{Shi2021OvercomingCF}                     &                      & 25.00$\pm$0.00          & 16.80$\pm$0.88          & 13.83$\pm$1.32          & 11.86$\pm$0.76          & 16.87$\pm$0.88          & -35.24$\pm$2.03                                                            & 16.49$\pm$0.00          & 13.51$\pm$0.80          & 9.98$\pm$0.56           & 9.08$\pm$0.61           & 12.27$\pm$0.58          & -49.18$\pm$0.80                                                            \\
FSLL\cite{Mazumder2021FewShotLL}                    &                      & 22.50$\pm$0.00          & 10.47$\pm$0.92          & 8.61$\pm$0.90           & 8.05$\pm$0.81           & 12.41$\pm$0.76          & -39.7$\pm$1.98                                                             & 25.78$\pm$0.00          & 12.21$\pm$1.73          & 9.58$\pm$1.86           & 7.85$\pm$0.61           & 13.86$\pm$1.31          & -47.59$\pm$1.42                                                            \\
ICARL\cite{Rebuffi2017iCaRLIC}                   &                      & 23.92$\pm$1.28          & 6.67$\pm$0.00           & 5.56$\pm$0.00           & 4.76$\pm$0.00           & 10.23$\pm$0.64          & -41.88$\pm$1.94                                                            & 16.56$\pm$0.43          & 5.41$\pm$0.00           & 4.34$\pm$0.00           & 3.60$\pm$0.00           & 7.48$\pm$0.22           & -53.97$\pm$0.59                                                            \\
Rebalance\cite{Hou2019LearningAU}               &                      & 12.75$\pm$0.44          & 9.73$\pm$0.39           & 8.39$\pm$0.18           & 7.10$\pm$0.20           & 9.49$\pm$0.32           & -42.62$\pm$1.86                                                            & 7.52$\pm$0.00           & 5.41$\pm$0.00           & 4.34$\pm$0.00           & 3.60$\pm$0.00           & 5.22$\pm$0.00           & -56.23$\pm$0.55                                                            \\
S3C\cite{Kalla2023S3CSS}                     &                      & 50.00$\pm$0.00          & 41.47$\pm$0.68          & 34.44$\pm$2.05          & 29.00$\pm$1.26          & 38.73$\pm$1.25          & -13.38$\pm$2.22                                                            & 49.27$\pm$0.00          & 33.87$\pm$0.72          & 27.29$\pm$1.14          & 23.19$\pm$0.85          & 33.41$\pm$0.80          & -28.04$\pm$0.97                                                            \\
SCN\cite{Ye2022BetterMB}                     &                      & 23.75$\pm$0.00          & 10.73$\pm$1.02          & 9.44$\pm$1.91           & 7.57$\pm$1.06           & 12.87$\pm$1.21          & -39.24$\pm$2.19                                                            & 36.33$\pm$0.00          & 11.92$\pm$1.24          & 11.14$\pm$0.97          & 11.70$\pm$0.89          & 17.77$\pm$0.90          & -43.68$\pm$1.05                                                            \\
\textcolor{orange}{ORCO\cite{ahmed2024orco}}                    &                      & 42.50$\pm$0.00          & 11.40$\pm$2.32          & 18.78$\pm$1.38          & 17.67$\pm$3.44          & 22.59$\pm$2.19          & -29.52$\pm$2.85                                                            & 31.94$\pm$0.00          & 15.36$\pm$1.39          & 10.04$\pm$1.96          & 10.16$\pm$1.94          & 16.88$\pm$1.54          & -44.57$\pm$1.64                                                            \\
\textbf{UNISA(Ours)}    &                      & \textbf{67.50$\pm$0.00} & \textbf{55.47$\pm$2.62} & \textbf{45.61$\pm$0.89} & \textbf{39.86$\pm$2.41} & \textbf{52.11$\pm$1.83} & \textbf{0.00$\pm$0.00}                                                     & \textbf{85.91$\pm$0.00} & \textbf{63.42$\pm$0.80} & \textbf{52.93$\pm$0.74} & \textbf{43.54$\pm$0.21} & \textbf{61.45$\pm$0.55} & \textbf{0.00$\pm$0.00}                                                     \\ \hline
Baseline\cite{Shi2021OvercomingCF}                & \multirow{9}{*}{3}   & 18.33$\pm$0.00          & 17.87$\pm$2.16          & 10.06$\pm$2.10          & 9.29$\pm$1.34           & 13.89$\pm$1.65          & -39.08$\pm$2.01                                                            & 18.06$\pm$0.00          & 12.01$\pm$0.80          & 9.31$\pm$0.43           & 7.91$\pm$0.71           & 11.82$\pm$0.58          & -48.71$\pm$0.69                                                            \\
F2M\cite{Shi2021OvercomingCF}                     &                      & 25.00$\pm$0.00          & 16.93$\pm$0.98          & 12.56$\pm$1.47          & 10.81$\pm$0.82          & 16.33$\pm$0.97          & -36.64$\pm$1.5                                                             & 16.49$\pm$0.00          & 13.17$\pm$1.13          & 10.17$\pm$1.22          & 9.28$\pm$0.72           & 12.28$\pm$0.91          & -48.25$\pm$0.99                                                            \\
FSLL\cite{Mazumder2021FewShotLL}                    &                      & 22.50$\pm$0.00          & 10.93$\pm$1.05          & 9.61$\pm$1.01           & 7.81$\pm$0.24           & 12.71$\pm$0.74          & -40.26$\pm$1.37                                                            & 25.78$\pm$0.00          & 13.08$\pm$1.07          & 9.80$\pm$1.76           & 9.04$\pm$1.09           & 14.43$\pm$1.17          & -46.10$\pm$1.23                                                            \\
ICARL\cite{Rebuffi2017iCaRLIC}                   &                      & 23.50$\pm$0.50          & 6.67$\pm$0.00           & 5.56$\pm$0.00           & 4.76$\pm$0.00           & 10.12$\pm$0.25          & -42.85$\pm$1.18                                                            & 16.62$\pm$0.37          & 5.41$\pm$0.00           & 4.34$\pm$0.00           & 3.60$\pm$0.00           & 7.49$\pm$0.19           & -53.04$\pm$0.42                                                            \\
Rebalance\cite{Hou2019LearningAU}               &                      & 12.50$\pm$0.23          & 9.80$\pm$0.44           & 8.56$\pm$0.18           & 7.14$\pm$0.19           & 9.50$\pm$0.28           & -43.47$\pm$1.18                                                            & 7.52$\pm$0.00           & 5.41$\pm$0.00           & 4.34$\pm$0.00           & 3.60$\pm$0.00           & 5.22$\pm$0.00           & -55.31$\pm$0.38                                                            \\
S3C\cite{Kalla2023S3CSS}                     &                      & 50.00$\pm$0.00          & 42.13$\pm$0.68          & 34.78$\pm$0.60          & 28.95$\pm$0.86          & 38.97$\pm$0.62          & -14.00$\pm$1.31                                                            & 49.27$\pm$0.00          & 32.79$\pm$1.20          & 26.45$\pm$0.86          & 22.56$\pm$0.59          & 32.77$\pm$0.79          & -27.76$\pm$0.88                                                            \\
SCN\cite{Ye2022BetterMB}                     &                      & 23.75$\pm$0.00          & 10.80$\pm$1.01          & 7.89$\pm$1.34           & 6.19$\pm$1.79           & 12.16$\pm$1.23          & -40.81$\pm$1.68                                                            & 36.33$\pm$0.00          & 11.38$\pm$1.66          & 9.93$\pm$1.16           & 10.04$\pm$1.18          & 16.92$\pm$1.17          & -43.61$\pm$1.23                                                            \\
\textcolor{orange}{ORCO\cite{ahmed2024orco}}                    &                      & 42.50$\pm$0.00          & 12.87$\pm$0.88          & 17.44$\pm$2.64          & 15.57$\pm$2.83          & 22.1$\pm$1.98           & -30.87$\pm$2.29                                                            & 31.94$\pm$0.00          & 12.22$\pm$1.62          & 9.41$\pm$1.40           & 10.62$\pm$0.69          & 16.05$\pm$1.12          & -44.48$\pm$1.18                                                            \\
\textbf{UNISA(Ours)}    &                      & \textbf{67.50$\pm$0.00} & \textbf{56.47$\pm$1.85} & \textbf{48.28$\pm$1.16} & \textbf{39.62$\pm$0.72} & \textbf{52.97$\pm$1.15} & \textbf{0.00$\pm$0.00}                                                     & \textbf{85.91$\pm$0.00} & \textbf{62.46$\pm$0.34} & \textbf{51.06$\pm$0.28} & \textbf{42.68$\pm$0.61} & \textbf{60.53$\pm$0.38} & \textbf{0.00$\pm$0.00}                                                     \\  \hline
Baseline\cite{Shi2021OvercomingCF}                & \multirow{9}{*}{2}   & 18.33$\pm$0.00          & 15.00$\pm$2.76          & 9.67$\pm$0.98           & 7.57$\pm$0.99           & 12.64$\pm$1.55          & -39.42$\pm$1.88                                                            & 18.06$\pm$0.00          & 10.75$\pm$0.85          & 9.30$\pm$1.12           & 8.12$\pm$0.50           & 11.56$\pm$0.75          & -49.28$\pm$0.87                                                            \\
F2M\cite{Shi2021OvercomingCF}                     &                      & 25.00$\pm$0.00          & 17.33$\pm$0.69          & 11.94$\pm$0.22          & 10.71$\pm$0.71          & 16.25$\pm$0.51          & -35.81$\pm$1.19                                                            & 16.49$\pm$0.00          & 13.33$\pm$0.79          & 10.33$\pm$0.38          & 9.04$\pm$0.41           & 12.3$\pm$0.48           & -48.54$\pm$0.66                                                            \\
FSLL\cite{Mazumder2021FewShotLL}                    &                      & 22.50$\pm$0.00          & 9.87$\pm$0.54           & 9.56$\pm$0.73           & 7.90$\pm$0.82           & 12.46$\pm$0.61          & -39.6$\pm$1.23                                                             & 25.78$\pm$0.00          & 11.83$\pm$0.90          & 9.78$\pm$0.88           & 7.69$\pm$1.70           & 13.77$\pm$1.06          & -47.07$\pm$1.15                                                            \\
ICARL\cite{Rebuffi2017iCaRLIC}                   &                      & 25.42$\pm$1.41          & 6.67$\pm$0.00           & 5.56$\pm$0.00           & 4.76$\pm$0.00           & 10.60$\pm$0.71          & -41.46$\pm$1.28                                                            & 16.58$\pm$0.46          & 5.41$\pm$0.00           & 4.34$\pm$0.00           & 3.60$\pm$0.00           & 7.48$\pm$0.23           & -53.36$\pm$0.51                                                            \\
Rebalance\cite{Hou2019LearningAU}               &                      & 12.50$\pm$0.57          & 10.13$\pm$0.23          & 8.50$\pm$0.19           & 7.14$\pm$0.13           & 9.57$\pm$0.33           & -42.49$\pm$1.12                                                            & 7.52$\pm$0.00           & 5.41$\pm$0.00           & 4.34$\pm$0.00           & 3.60$\pm$0.00           & 5.22$\pm$0.00           & -55.62$\pm$0.45                                                            \\
S3C\cite{Kalla2023S3CSS}                     &                      & 50.00$\pm$0.00          & 40.60$\pm$1.19          & 35.22$\pm$1.00          & 28.62$\pm$1.65          & 38.61$\pm$1.13          & -13.45$\pm$1.56                                                            & 49.27$\pm$0.00          & 33.69$\pm$0.30          & 26.20$\pm$0.98          & 22.81$\pm$0.65          & 32.99$\pm$0.61          & -27.85$\pm$0.76                                                            \\
SCN\cite{Ye2022BetterMB}                     &                      & 23.75$\pm$0.00          & 8.20$\pm$1.45           & 7.39$\pm$1.68           & 6.67$\pm$1.02           & 11.50$\pm$1.22          & -40.56$\pm$1.62                                                            & 36.33$\pm$0.00          & 9.49$\pm$1.41           & 7.10$\pm$0.92           & 8.79$\pm$1.10           & 15.43$\pm$1.01          & -45.41$\pm$1.11                                                            \\
\textcolor{orange}{ORCO\cite{ahmed2024orco}}                    &                      & 42.50$\pm$0.00          & 11.07$\pm$0.91          & 16.22$\pm$1.51          & 15.81$\pm$2.19          & 21.40$\pm$1.41          & -30.66$\pm$1.77                                                            & 31.94$\pm$0.00          & 11.31$\pm$2.32          & 8.61$\pm$1.50           & 10.52$\pm$3.37          & 15.6$\pm$2.18           & -45.24$\pm$2.23                                                            \\
\textbf{UNISA(Ours)}    &                      & \textbf{67.50$\pm$0.00} & \textbf{55.00$\pm$1.21} & \textbf{46.39$\pm$1.38} & \textbf{39.33$\pm$1.11} & \textbf{52.06$\pm$1.07} & \textbf{0.00$\pm$0.00}                                                     & \textbf{85.91$\pm$0.00} & \textbf{63.20$\pm$0.39} & \textbf{51.05$\pm$0.69} & \textbf{43.20$\pm$0.41} & \textbf{60.84$\pm$0.45} & \textbf{0.00$\pm$0.00}                                                     \\  \hline
Baseline\cite{Shi2021OvercomingCF}               & \multirow{9}{*}{1}   & 18.33$\pm$0.00          & 12.33$\pm$2.89          & 9.67$\pm$0.93           & 8.19$\pm$0.67           & 12.13$\pm$1.55          & -38.39$\pm$2.32                                                            & 18.06$\pm$0.00          & 11.53$\pm$0.80          & 9.59$\pm$0.35           & 8.13$\pm$0.15           & 11.83$\pm$0.44          & -48.10$\pm$0.79                                                            \\
F2M\cite{Shi2021OvercomingCF}                     &                      & 25.00$\pm$0.00          & 18.07$\pm$3.34          & 12.28$\pm$1.04          & 10.90$\pm$1.21          & 16.56$\pm$1.85          & -33.96$\pm$2.53                                                            & 16.49$\pm$0.00          & 13.68$\pm$0.94          & 9.95$\pm$0.45           & 8.88$\pm$0.48           & 12.25$\pm$0.57          & -47.68$\pm$0.87                                                            \\
FSLL\cite{Mazumder2021FewShotLL}                    &                      & 22.50$\pm$0.00          & 10.40$\pm$1.02          & 8.06$\pm$0.74           & 7.05$\pm$0.74           & 12.00$\pm$0.73          & -38.52$\pm$1.87                                                            & 25.78$\pm$0.00          & 11.04$\pm$1.00          & 9.84$\pm$0.68           & 8.90$\pm$1.19           & 13.89$\pm$0.85          & -46.04$\pm$1.08                                                            \\
ICARL\cite{Rebuffi2017iCaRLIC}                   &                      & 23.92$\pm$0.85          & 6.67$\pm$0.00           & 5.56$\pm$0.00           & 4.76$\pm$0.00           & 10.23$\pm$0.43          & -40.29$\pm$1.77                                                            & 16.47$\pm$0.33          & 5.41$\pm$0.00           & 4.34$\pm$0.00           & 3.60$\pm$0.00           & 7.46$\pm$0.17           & -52.47$\pm$0.68                                                            \\
Rebalance\cite{Hou2019LearningAU}               &                      & 12.58$\pm$0.15          & 9.87$\pm$0.30           & 8.50$\pm$0.12           & 7.29$\pm$0.21           & 9.56$\pm$0.21           & -40.96$\pm$1.73                                                            & 7.52$\pm$0.00           & 5.41$\pm$0.00           & 4.34$\pm$0.00           & 3.60$\pm$0.00           & 5.22$\pm$0.00           & -54.71$\pm$0.66                                                            \\
S3C\cite{Kalla2023S3CSS}                     &                      & 50.00$\pm$0.00          & 40.33$\pm$1.00          & 34.89$\pm$1.26          & 27.67$\pm$1.24          & 38.22$\pm$1.02          & -12.30$\pm$2.00                                                            & 49.27$\pm$0.00          & 32.00$\pm$1.26          & 24.80$\pm$0.29          & 20.56$\pm$0.70          & 31.66$\pm$0.74          & -28.27$\pm$0.99                                                            \\
SCN\cite{Ye2022BetterMB}                     &                      & 23.75$\pm$0.00          & 7.40$\pm$1.58           & 6.61$\pm$1.70           & 5.76$\pm$0.79           & 10.88$\pm$1.23          & -39.64$\pm$2.11                                                            & 36.33$\pm$0.00          & 8.48$\pm$1.02           & 6.43$\pm$0.51           & 7.41$\pm$0.61           & 14.66$\pm$0.65          & -45.27$\pm$0.93                                                            \\
\textcolor{orange}{ORCO\cite{ahmed2024orco}}                    &                      & 42.50$\pm$0.00          & 12.87$\pm$1.04          & 14.61$\pm$1.91          & 8.86$\pm$1.63           & 19.71$\pm$1.36          & -30.81$\pm$2.19                                                            & 31.94$\pm$0.00          & 13.50$\pm$1.16          & 8.53$\pm$0.64           & 7.58$\pm$0.82           & 15.39$\pm$0.78          & -44.14$\pm$1.02                                                            \\
\textbf{UNISA(Ours)}    &                      & \textbf{67.50$\pm$0.00} & \textbf{55.33$\pm$0.58} & \textbf{43.00$\pm$2.56} & \textbf{36.24$\pm$2.22} & \textbf{50.52$\pm$1.72} & \textbf{0.00$\pm$0.00}                                                     & \textbf{85.91$\pm$0.00} & \textbf{62.31$\pm$0.47} & \textbf{50.19$\pm$0.32} & \textbf{41.31$\pm$1.20} & \textbf{59.93$\pm$0.66} & \textbf{0.00$\pm$0.00}  \\ \hline                                                              
\end{tabular}
\caption{\textcolor{blue}{Classification accuracy on split UCMerced and AID dataset with 1-5 shot setting
averaged across 5 times run.}}
\label{tab:ucmerced_and_aid}
\end{table*}

\begin{table*}[]
\centering
\setlength{\tabcolsep}{0.21em}
\scriptsize
\begin{tabular}{l|c|cccccc|cccccc}
\hline
\multirow{3}{*}{Method} & \multirow{3}{*}{\#S} & \multicolumn{6}{c}{Split RSICB256}                                                                                                                     & \multicolumn{6}{c}{Split RESISC45}                                                                                                                          \\ \hline
                        &                      & \multicolumn{4}{c}{Session}                           & \multirow{2}{*}{Avg} & \multirow{2}{*}{\begin{tabular}[c]{@{}c@{}}Gap to\\ UNISA\end{tabular}} & \multicolumn{4}{c}{Session}                           & \multirow{2}{*}{Avg} & \multirow{2}{*}{\begin{tabular}[c]{@{}c@{}}Gap to\\ UNISA\end{tabular}} \\ \cline{3-6} \cline{9-12}
                        &                      & 1           & 2           & 3           & 4           &                      &                                                                         & 1           & 2           & 3           & 4           &                      &                                                                         \\ \hline
Baseline\cite{Shi2021OvercomingCF}                & \multirow{9}{*}{5}   & 41.71$\pm$0.00          & 33.18$\pm$0.47          & 28.39$\pm$1.07          & 28.50$\pm$0.61          & 32.95$\pm$0.66          & -44.12$\pm$0.98                                                            & 16.93$\pm$0.00          & 11.24$\pm$0.00          & 10.21$\pm$0.00          & 8.63$\pm$0.00           & 11.75$\pm$0.00          & -54.66$\pm$0.00                                                            \\
F2M\cite{Shi2021OvercomingCF}                     &                      & 21.78$\pm$0.00          & 18.19$\pm$0.73          & 16.25$\pm$0.63          & 14.64$\pm$1.16          & 17.72$\pm$0.75          & -59.35$\pm$1.04                                                            & 16.52$\pm$0.00          & 13.00$\pm$0.00          & 10.92$\pm$0.00          & 9.33$\pm$0.00           & 12.44$\pm$0.00          & -53.97$\pm$0.00                                                            \\
FSLL\cite{Mazumder2021FewShotLL}                    &                      & 42.21$\pm$0.00          & 24.18$\pm$2.05          & 21.02$\pm$0.55          & 23.28$\pm$0.90          & 27.67$\pm$1.15          & -49.40$\pm$1.36                                                            & 16.70$\pm$0.00          & 11.43$\pm$0.00          & 8.46$\pm$0.00           & 7.05$\pm$0.00           & 10.91$\pm$0.00          & -55.50$\pm$0.00                                                            \\
ICARL\cite{Rebuffi2017iCaRLIC}                   &                      & 31.91$\pm$0.62          & 2.23$\pm$0.00           & 1.76$\pm$0.00           & 1.43$\pm$0.00           & 9.33$\pm$0.31           & -67.74$\pm$0.78                                                            & 16.92$\pm$0.18          & 3.23$\pm$0.00           & 2.63$\pm$0.00           & 2.22$\pm$0.00           & 6.25$\pm$0.09           & -60.16$\pm$0.09                                                            \\
Rebalance\cite{Hou2019LearningAU}               &                      & 6.54$\pm$0.06           & 5.33$\pm$0.12           & 4.25$\pm$0.03           & 3.45$\pm$0.03           & 4.89$\pm$0.07           & -72.18$\pm$0.72                                                            & 4.32$\pm$0.02           & 3.31$\pm$0.01           & 2.70$\pm$0.01           & 2.28$\pm$0.01           & 3.15$\pm$0.01           & -63.26$\pm$0.01                                                            \\
S3C\cite{Kalla2023S3CSS}                     &                      & 82.74$\pm$0.00          & 69.33$\pm$0.41          & 62.10$\pm$0.71          & 57.23$\pm$0.21          & 67.85$\pm$0.42          & -9.22$\pm$0.83                                                             & 47.11$\pm$0.00          & 36.71$\pm$0.64          & 30.41$\pm$0.21          & 26.04$\pm$0.24          & 35.07$\pm$0.36          & -31.34$\pm$0.36                                                            \\
SCN\cite{Ye2022BetterMB}                     &                      & 33.60$\pm$0.00          & 8.32$\pm$0.42           & 17.42$\pm$0.59          & 14.18$\pm$1.30          & 18.38$\pm$0.74          & -58.69$\pm$1.03                                                            & 14.61$\pm$0.00          & 5.37$\pm$0.46           & 6.48$\pm$0.38           & 6.63$\pm$0.56           & 8.27$\pm$0.41           & -58.14$\pm$0.41                                                            \\
\textcolor{orange}{ORCO\cite{ahmed2024orco}}                    &                      & 40.83$\pm$0.00          & 31.67$\pm$0.00          & 15.66$\pm$0.00          & 21.12$\pm$0.00          & 27.10$\pm$0.00          & -49.97$\pm$1.16                                                            & 34.35$\pm$0.00          & 8.86$\pm$0.94           & 10.67$\pm$1.09          & 10.06$\pm$1.12          & 15.99$\pm$0.91          & -50.43$\pm$0.91                                                            \\
\textbf{UNISA(Ours)}    &                      & \textbf{95.41$\pm$0.00} & \textbf{80.36$\pm$0.83} & \textbf{70.08$\pm$0.69} & \textbf{62.43$\pm$0.94} & \textbf{77.07$\pm$0.72} & \textbf{0.00$\pm$0.00}                                                     & \textbf{88.51$\pm$0.00} & \textbf{69.56$\pm$0.21} & \textbf{57.93$\pm$0.18} & \textbf{49.62$\pm$0.71} & \textbf{66.41$\pm$0.38} & \textbf{0.00$\pm$0.00}                                                     \\ \hline
Baseline\cite{Shi2021OvercomingCF}                & \multirow{9}{*}{4}   & 41.71$\pm$0.00          & 33.00$\pm$0.12          & 28.41$\pm$0.75          & 28.13$\pm$0.68          & 32.81$\pm$0.51          & -43.41$\pm$1.05                                                            & 16.93$\pm$0.00          & 11.89$\pm$0.00          & 10.98$\pm$0.00          & 9.10$\pm$0.00           & 12.23$\pm$0.00          & -53.83$\pm$0.00                                                            \\
F2M\cite{Shi2021OvercomingCF}                     &                      & 21.78$\pm$0.00          & 18.19$\pm$0.65          & 16.52$\pm$0.36          & 15.32$\pm$1.16          & 17.95$\pm$0.69          & -58.27$\pm$1.15                                                            & 16.01$\pm$0.00          & 13.29$\pm$0.00          & 11.45$\pm$0.00          & 9.08$\pm$0.00           & 12.46$\pm$0.00          & -53.59$\pm$0.00                                                            \\
FSLL\cite{Mazumder2021FewShotLL}                    &                      & 42.21$\pm$0.00          & 24.18$\pm$2.05          & 21.02$\pm$0.55          & 23.28$\pm$0.90          & 27.67$\pm$1.15          & -48.55$\pm$1.47                                                            & 16.55$\pm$0.00          & 11.29$\pm$0.00          & 7.80$\pm$0.00           & 6.63$\pm$0.00           & 10.57$\pm$0.00          & -55.48$\pm$0.00                                                            \\
ICARL\cite{Rebuffi2017iCaRLIC}                   &                      & 32.42$\pm$0.49          & 2.23$\pm$0.00           & 1.76$\pm$0.00           & 1.43$\pm$0.00           & 9.46$\pm$0.25           & -66.76$\pm$0.95                                                            & 17.05$\pm$0.17          & 3.23$\pm$0.00           & 2.63$\pm$0.00           & 2.22$\pm$0.00           & 6.28$\pm$0.09           & -59.77$\pm$0.09                                                            \\
Rebalance\cite{Hou2019LearningAU}               &                      & 6.54$\pm$0.07           & 5.42$\pm$0.08           & 4.20$\pm$0.07           & 3.43$\pm$0.04           & 4.90$\pm$0.07           & -71.32$\pm$0.92                                                            & 4.29$\pm$0.02           & 3.32$\pm$0.01           & 2.71$\pm$0.00           & 2.29$\pm$0.00           & 3.15$\pm$0.01           & -62.90$\pm$0.01                                                            \\
S3C\cite{Kalla2023S3CSS}                     &                      & 82.74$\pm$0.00          & 68.81$\pm$0.32          & 61.30$\pm$0.77          & 56.00$\pm$0.42          & 67.21$\pm$0.47          & -9.01$\pm$1.03                                                             & 47.11$\pm$0.00          & 36.82$\pm$0.47          & 30.50$\pm$0.54          & 25.90$\pm$0.52          & 35.08$\pm$0.44          & -30.97$\pm$0.44                                                            \\
SCN\cite{Ye2022BetterMB}                     &                      & 33.60$\pm$0.00          & 8.74$\pm$0.66           & 16.33$\pm$1.08          & 13.31$\pm$1.03          & 18.00$\pm$0.82          & -58.22$\pm$1.23                                                            & 14.61$\pm$0.00          & 5.29$\pm$0.43           & 6.64$\pm$0.62           & 6.04$\pm$0.48           & 8.15$\pm$0.45           & -57.91$\pm$0.45                                                            \\
\textcolor{orange}{ORCO\cite{ahmed2024orco}}                    &                      & 40.83$\pm$0.00          & 31.57$\pm$0.00          & 26.42$\pm$0.00          & 25.45$\pm$0.00          & 28.21$\pm$0.00          & -48.01$\pm$1.38                                                            & 34.35$\pm$0.00          & 10.12$\pm$0.76          & 8.89$\pm$1.82           & 9.38$\pm$0.57           & 15.69$\pm$1.03          & -50.37$\pm$1.03                                                            \\
\textbf{UNISA(Ours)}    &                      & \textbf{95.41$\pm$0.00} & \textbf{80.15$\pm$0.60} & \textbf{68.69$\pm$1.08} & \textbf{60.63$\pm$1.36} & \textbf{76.22$\pm$0.92} & \textbf{0.00$\pm$0.00}                                                     & \textbf{88.51$\pm$0.00} & \textbf{69.36$\pm$0.40} & \textbf{57.42$\pm$0.46} & \textbf{48.90$\pm$0.45} & \textbf{66.05$\pm$0.38} & \textbf{0.00$\pm$0.00}                                                     \\ \hline
Baseline\cite{Shi2021OvercomingCF}                & \multirow{9}{*}{3}   & 41.71$\pm$0.00          & 32.78$\pm$0.33          & 27.89$\pm$1.26          & 27.35$\pm$1.46          & 32.43$\pm$0.98          & -42.34$\pm$1.23                                                            & 17.05$\pm$0.00          & 11.96$\pm$0.00          & 10.60$\pm$0.00          & 8.86$\pm$0.00           & 12.12$\pm$0.00          & -53.29$\pm$0.00                                                            \\
F2M\cite{Shi2021OvercomingCF}                     &                      & 21.78$\pm$0.00          & 18.26$\pm$0.94          & 16.68$\pm$0.75          & 14.84$\pm$1.23          & 17.89$\pm$0.86          & -56.88$\pm$1.13                                                            & 16.07$\pm$0.00          & 12.10$\pm$0.00          & 11.13$\pm$0.00          & 9.79$\pm$0.00           & 12.27$\pm$0.00          & -53.14$\pm$0.00                                                            \\
FSLL\cite{Mazumder2021FewShotLL}                    &                      & 42.21$\pm$0.00          & 22.56$\pm$2.15          & 19.92$\pm$1.13          & 22.14$\pm$1.30          & 26.71$\pm$1.38          & -48.06$\pm$1.57                                                            & 16.34$\pm$0.00          & 10.51$\pm$0.00          & 8.89$\pm$0.00           & 7.44$\pm$0.00           & 10.80$\pm$0.00          & -54.62$\pm$0.00                                                            \\
ICARL\cite{Rebuffi2017iCaRLIC}                   &                      & 32.52$\pm$0.47          & 2.23$\pm$0.00           & 1.76$\pm$0.00           & 1.43$\pm$0.00           & 9.49$\pm$0.24           & -65.28$\pm$0.78                                                            & 16.98$\pm$0.09          & 3.23$\pm$0.00           & 2.63$\pm$0.00           & 2.22$\pm$0.00           & 6.27$\pm$0.05           & -59.15$\pm$0.05                                                            \\
Rebalance\cite{Hou2019LearningAU}               &                      & 6.58$\pm$0.13           & 5.35$\pm$0.04           & 4.22$\pm$0.05           & 3.51$\pm$0.03           & 4.92$\pm$0.07           & -69.85$\pm$0.74                                                            & 4.29$\pm$0.02           & 3.33$\pm$0.01           & 2.70$\pm$0.01           & 2.29$\pm$0.01           & 3.15$\pm$0.01           & -62.26$\pm$0.01                                                            \\
S3C\cite{Kalla2023S3CSS}                     &                      & 82.74$\pm$0.00          & 69.35$\pm$0.49          & 61.02$\pm$0.90          & 55.13$\pm$0.72          & 67.06$\pm$0.63          & -7.71$\pm$0.97                                                             & 47.11$\pm$0.00          & 36.13$\pm$0.62          & 29.92$\pm$0.55          & 25.06$\pm$0.40          & 34.56$\pm$0.46          & -30.86$\pm$0.46                                                            \\
SCN\cite{Ye2022BetterMB}                     &                      & 33.60$\pm$0.00          & 8.65$\pm$0.30           & 15.10$\pm$1.18          & 12.09$\pm$0.87          & 17.36$\pm$0.75          & -57.41$\pm$1.05                                                            & 14.61$\pm$0.00          & 5.40$\pm$0.13           & 5.89$\pm$0.52           & 6.02$\pm$0.94           & 7.98$\pm$0.54           & -57.43$\pm$0.54                                                            \\
\textcolor{orange}{ORCO\cite{ahmed2024orco}}                    &                      & 40.83$\pm$0.00          & 27.70$\pm$0.00          & 25.21$\pm$0.00          & 27.39$\pm$0.00          & 27.61$\pm$0.00          & -47.16$\pm$1.35                                                            & 34.35$\pm$0.00          & 8.69$\pm$1.02           & 9.00$\pm$1.19           & 7.99$\pm$1.62           & 15.01$\pm$1.13          & -50.40$\pm$1.13                                                            \\
\textbf{UNISA(Ours)}    &                      & \textbf{95.41$\pm$0.00} & \textbf{78.77$\pm$0.79} & \textbf{66.31$\pm$0.61} & \textbf{58.58$\pm$1.09} & \textbf{74.77$\pm$0.74} & \textbf{0.00$\pm$0.00}                                                     & \textbf{88.51$\pm$0.00} & \textbf{68.82$\pm$0.56} & \textbf{56.29$\pm$0.81} & \textbf{48.03$\pm$0.74} & \textbf{65.41$\pm$0.62} & \textbf{0.00$\pm$0.00}                                                     \\ \hline
Baselin\cite{Shi2021OvercomingCF}                & \multirow{9}{*}{2}   & 41.71$\pm$0.00          & 32.49$\pm$0.86          & 26.67$\pm$0.68          & 25.69$\pm$1.36          & 31.64$\pm$0.87          & -42.43$\pm$1.26                                                            & 16.73$\pm$0.00          & 11.24$\pm$0.00          & 9.94$\pm$0.00           & 7.98$\pm$0.00           & 11.47$\pm$0.00          & -53.95$\pm$0.00                                                            \\
F2M\cite{Shi2021OvercomingCF}                     &                      & 21.78$\pm$0.00          & 17.65$\pm$0.66          & 15.82$\pm$1.05          & 14.85$\pm$1.62          & 17.53$\pm$1.02          & -56.54$\pm$1.37                                                            & 15.48$\pm$0.00          & 11.77$\pm$0.00          & 10.28$\pm$0.00          & 8.94$\pm$0.00           & 11.62$\pm$0.00          & -53.80$\pm$0.00                                                            \\
FSLL\cite{Mazumder2021FewShotLL}                    &                      & 42.21$\pm$0.00          & 22.03$\pm$1.86          & 19.23$\pm$2.01          & 21.73$\pm$1.14          & 26.3$\pm$1.48           & -47.77$\pm$1.74                                                            & 16.93$\pm$0.00          & 9.65$\pm$0.00           & 8.14$\pm$0.00           & 6.13$\pm$0.00           & 10.21$\pm$0.00          & -55.21$\pm$0.00                                                            \\
ICARL\cite{Rebuffi2017iCaRLIC}                   &                      & 32.28$\pm$0.60          & 2.23$\pm$0.00           & 1.76$\pm$0.00           & 1.43$\pm$0.00           & 9.43$\pm$0.30           & -64.64$\pm$0.96                                                            & 17.10$\pm$0.12          & 3.23$\pm$0.00           & 2.63$\pm$0.00           & 2.22$\pm$0.00           & 6.30$\pm$0.06           & -59.13$\pm$0.06                                                            \\
Rebalance\cite{Hou2019LearningAU}               &                      & 6.57$\pm$0.11           & 5.40$\pm$0.06           & 4.20$\pm$0.08           & 3.44$\pm$0.05           & 4.90$\pm$0.08           & -69.17$\pm$0.91                                                            & 4.30$\pm$0.01           & 3.32$\pm$0.01           & 2.71$\pm$0.01           & 2.29$\pm$0.01           & 3.16$\pm$0.01           & -62.27$\pm$0.01                                                            \\
S3C\cite{Kalla2023S3CSS}                     &                      & 82.74$\pm$0.00          & 68.73$\pm$0.71          & 58.97$\pm$1.43          & 53.32$\pm$1.02          & 65.94$\pm$0.95          & -8.13$\pm$1.32                                                             & 47.11$\pm$0.00          & 36.09$\pm$0.58          & 29.11$\pm$0.16          & 24.63$\pm$0.47          & 34.24$\pm$0.38          & -31.19$\pm$0.38                                                            \\
SCN\cite{Ye2022BetterMB}                     &                      & 33.60$\pm$0.00          & 7.15$\pm$1.61           & 13.62$\pm$1.20          & 10.77$\pm$1.22          & 16.29$\pm$1.17          & -57.78$\pm$1.48                                                            & 14.61$\pm$0.00          & 5.25$\pm$0.55           & 4.83$\pm$0.63           & 5.03$\pm$0.57           & 7.43$\pm$0.51           & -57.99$\pm$0.51                                                            \\
\textcolor{orange}{ORCO\cite{ahmed2024orco}}                    &                      & 40.83$\pm$0.00          & 24.84$\pm$0.00          & 23.58$\pm$0.00          & 5.38$\pm$0.00           & 25.99$\pm$0.,00         & -48.08$\pm$1.83                                                            & 34.35$\pm$0.00          & 8.58$\pm$1.27           & 7.88$\pm$2.35           & 8.20$\pm$1.74           & 14.75$\pm$1.59          & -50.67$\pm$1.59                                                            \\
\textbf{UNISA(Ours)}    &                      & \textbf{95.41$\pm$0.00} & \textbf{79.19$\pm$1.00} & \textbf{65.19$\pm$0.91} & \textbf{56.49$\pm$1.21} & \textbf{74.07$\pm$0.91} & \textbf{0.00$\pm$0.00}                                                     & \textbf{88.51$\pm$0.00} & \textbf{68.53$\pm$0.66} & \textbf{56.65$\pm$0.39} & \textbf{47.99$\pm$0.65} & \textbf{65.42$\pm$0.50} & \textbf{0.00$\pm$0.00}                                                     \\ \hline
Baseline\cite{Shi2021OvercomingCF}                & \multirow{9}{*}{1}   & 41.71$\pm$0.00          & 32.50$\pm$0.62          & 26.13$\pm$1.18          & 24.40$\pm$0.73          & 31.19$\pm$0.76          & -40.68$\pm$1.21                                                            & 16.61$\pm$0.00          & 12.47$\pm$0.00          & 10.58$\pm$0.00          & 7.94$\pm$0.00           & 11.90$\pm$0.00          & -52.99$\pm$0.00                                                            \\
F2M\cite{Shi2021OvercomingCF}                     &                      & 21.78$\pm$0.00          & 16.33$\pm$1.44          & 15.54$\pm$0.79          & 13.51$\pm$0.74          & 16.79$\pm$0.90          & -55.08$\pm$1.30                                                            & 16.76$\pm$0.00          & 11.82$\pm$0.00          & 9.42$\pm$0.00           & 8.16$\pm$0.00           & 11.54$\pm$0.00          & -53.35$\pm$0.00                                                            \\
FSLL\cite{Mazumder2021FewShotLL}                    &                      & 42.21$\pm$0.00          & 19.69$\pm$1.78          & 18.07$\pm$1.96          & 19.78$\pm$2.47          & 24.94$\pm$1.81          & -46.93$\pm$2.04                                                            & 16.37$\pm$0.00          & 8.80$\pm$0.00           & 7.78$\pm$0.00           & 4.86$\pm$0.00           & 9.45$\pm$0.00           & -55.44$\pm$0.00                                                            \\
ICARL\cite{Rebuffi2017iCaRLIC}                   &                      & 32.76$\pm$0.36          & 2.23$\pm$0.00           & 1.76$\pm$0.00           & 1.43$\pm$0.00           & 9.55$\pm$0.18           & -62.32$\pm$0.96                                                            & 17.17$\pm$0.28          & 3.23$\pm$0.00           & 2.63$\pm$0.00           & 2.22$\pm$0.00           & 6.31$\pm$0.14           & -58.58$\pm$0.14                                                            \\
Rebalance\cite{Hou2019LearningAU}               &                      & 6.56$\pm$0.08           & 5.38$\pm$0.04           & 4.21$\pm$0.05           & 3.47$\pm$0.05           & 4.91$\pm$0.06           & -66.96$\pm$0.94                                                            & 4.30$\pm$0.01           & 3.33$\pm$0.02           & 2.72$\pm$0.01           & 2.29$\pm$0.01           & 3.16$\pm$0.01           & -61.73$\pm$0.01                                                            \\
S3C\cite{Kalla2023S3CSS}                     &                      & 82.74$\pm$0.00          & 67.37$\pm$0.77          & 58.27$\pm$0.99          & 51.83$\pm$0.71          & 65.05$\pm$0.72          & -6.82$\pm$1.18                                                             & 47.11$\pm$0.00          & 35.65$\pm$1.07          & 27.97$\pm$0.75          & 22.93$\pm$0.82          & 33.42$\pm$0.77          & -31.48$\pm$0.77                                                            \\
SCN\cite{Ye2022BetterMB}                     &                      & 33.60$\pm$0.00          & 5.43$\pm$0.80           & 8.56$\pm$2.04           & 8.31$\pm$1.83           & 13.98$\pm$1.43          & -57.89$\pm$1.71                                                            & 14.61$\pm$0.00          & 4.63$\pm$0.45           & 4.17$\pm$0.98           & 3.85$\pm$0.42           & 6.82$\pm$0.58           & -58.08$\pm$0.58                                                            \\
\textcolor{orange}{ORCO\cite{ahmed2024orco}}                    &                      & 40.83$\pm$0.00          & 10.72$\pm$0.00          & 13.61$\pm$0.00          & 14.79$\pm$0.00          & 20.55$\pm$0.00          & -51.32$\pm$1.22                                                            & 34.35$\pm$0.00          & 7.81$\pm$1.29           & 5.97$\pm$0.76           & 5.95$\pm$0.34           & 13.52$\pm$0.77          & -51.37$\pm$0.77                                                            \\
\textbf{UNISA(Ours)}    &                      & \textbf{95.41$\pm$0.00} & \textbf{77.68$\pm$0.71} & \textbf{62.37$\pm$1.12} & \textbf{52.02$\pm$1.32} & \textbf{71.87$\pm$0.94} & \textbf{0.00$\pm$0.00}                                                     & \textbf{88.51$\pm$0.00} & \textbf{68.26$\pm$0.47} & \textbf{55.68$\pm$0.57} & \textbf{47.09$\pm$0.57} & \textbf{64.89$\pm$0.47} & \textbf{0.00$\pm$0.00}   \\ \hline                                                              
\end{tabular}
\caption{\textcolor{blue}{Classification accuracy on split RSICB256 and split RESISC45 dataset with 1-5 shot setting
averaged across 5 times run.}}
\label{tab:rsicb256_and_resisc45}
\end{table*}

\begin{table*}[]
\setlength{\tabcolsep}{0.21em}
\scriptsize
\centering
\begin{tabular}{l|c|cccccc|cccccc}
\hline
\multirow{3}{*}{Method} & \multirow{3}{*}{\#S} & \multicolumn{6}{c}{Split EuroSAT}                                                                                                                     & \multicolumn{6}{c}{Split HyRank}                                                                                                                          \\ \hline
                        &                      & \multicolumn{4}{c}{Session}                           & \multirow{2}{*}{Avg} & \multirow{2}{*}{\begin{tabular}[c]{@{}c@{}}Gap to\\ UNISA\end{tabular}} & \multicolumn{4}{c}{Session}                           & \multirow{2}{*}{Avg} & \multirow{2}{*}{\begin{tabular}[c]{@{}c@{}}Gap to\\ UNISA\end{tabular}} \\ \cline{3-6} \cline{9-12}
                        &                      & 1           & 2           & 3           & 4           &                      &                                                                         & 1           & 2           & 3           & 4           &                      &                                                                         \\ \hline
Baseline\cite{Shi2021OvercomingCF}                & \multirow{9}{*}{5}   & 53.39$\pm$0.00          & 46.77$\pm$0.85          & 34.66$\pm$0.86          & 27.61$\pm$0.97          & 40.61$\pm$0.78          & -33.04$\pm$1.95                                                            & 33.42$\pm$0.00          & 26.56$\pm$3.59          & 21.51$\pm$2.33          & 23.37$\pm$2.03          & 26.22$\pm$2.37          & -17.48$\pm$2.65                                                            \\
F2M\cite{Shi2021OvercomingCF}                     &                      & 56.22$\pm$0.00          & 43.12$\pm$1.10          & 33.91$\pm$1.61          & 27.53$\pm$1.52          & 40.20$\pm$1.24          & -33.45$\pm$2.18                                                            & 16.79$\pm$0.00          & 10.27$\pm$2.44          & 9.26$\pm$2.36           & 5.43$\pm$1.15           & 10.44$\pm$1.79          & -33.26$\pm$2.14                                                            \\
FSLL\cite{Mazumder2021FewShotLL}                    &                      & 57.78$\pm$0.00          & 30.73$\pm$2.26          & 21.82$\pm$2.22          & 23.14$\pm$2.44          & 33.37$\pm$2.00          & -40.28$\pm$2.68                                                            & 35.27$\pm$0.00          & 26.12$\pm$1.46          & 21.54$\pm$2.10          & 23.91$\pm$0.53          & 26.71$\pm$1.31          & -16.99$\pm$1.76                                                            \\
ICARL\cite{Rebuffi2017iCaRLIC}                   &                      & 36.48$\pm$0.15          & 30.58$\pm$1.99          & 23.70$\pm$2.61          & 20.54$\pm$1.72          & 27.83$\pm$1.85          & -45.82$\pm$2.57                                                            & 26.78$\pm$1.03          & 31.29$\pm$0.00          & 29.44$\pm$0.00          & 29.06$\pm$0.00          & 29.14$\pm$0.52          & -14.56$\pm$1.29                                                            \\
Rebalance\cite{Hou2019LearningAU}               &                      & 26.09$\pm$0.00          & 18.75$\pm$0.00          & 13.95$\pm$0.00          & 11.11$\pm$0.00          & 17.48$\pm$0.00          & -56.17$\pm$1.79                                                            & 34.31$\pm$0.00          & 31.29$\pm$0.00          & 29.44$\pm$0.00          & 29.06$\pm$0.00          & 31.03$\pm$0.00          & -12.67$\pm$1.18                                                            \\
S3C\cite{Kalla2023S3CSS}                     &                      & 84.26$\pm$0.00          & 65.87$\pm$0.28          & 52.31$\pm$1.62          & 43.53$\pm$1.16          & 61.49$\pm$1.01          & -12.16$\pm$2.06                                                            & 49.86$\pm$0.00          & 34.31$\pm$0.83          & 30.71$\pm$0.60          & 28.76$\pm$1.03          & 35.91$\pm$0.73          & -7.79$\pm$1.39                                                             \\
SCN\cite{Ye2022BetterMB}                     &                      & 67.35$\pm$0.00          & 27.52$\pm$1.07          & 18.21$\pm$3.61          & 16.66$\pm$2.15          & 32.44$\pm$2.17          & -41.21$\pm$2.81                                                            & 47.48$\pm$0.00          & 8.74$\pm$0.06           & 8.16$\pm$1.11           & 5.10$\pm$1.56           & 17.37$\pm$0.96          & -26.33$\pm$1.52                                                            \\
\textcolor{orange}{ORCO\cite{ahmed2024orco}}                   &                      & 71.70$\pm$0.00          & 27.28$\pm$0.23          & 20.66$\pm$3.41          & 20.57$\pm$3.53          & 35.05$\pm$2.46          & -38.60$\pm$3.04                                                            & 36.37$\pm$0.00          & 7.86$\pm$0.59           & 4.65$\pm$1.36           & 1.70$\pm$0.84           & 12.65$\pm$0.85          & -31.05$\pm$1.45                                                            \\
\textbf{UNISA(Ours)}    &                      & \textbf{96.70$\pm$0.00} & \textbf{79.06$\pm$1.71} & \textbf{64.95$\pm$2.45} & \textbf{53.89$\pm$1.98} & \textbf{73.65$\pm$1.79} & \textbf{0.00$\pm$0.00}                                                     & \textbf{52.75$\pm$0.00} & \textbf{44.56$\pm$0.51} & \textbf{39.84$\pm$1.39} & \textbf{37.64$\pm$1.83} & \textbf{43.70$\pm$1.18} & \textbf{0.00$\pm$0.00}                                                     \\ \hline
Baseline\cite{Shi2021OvercomingCF}                & \multirow{9}{*}{4}   & 53.39$\pm$0.00          & 46.99$\pm$0.45          & 34.20$\pm$0.67          & 27.46$\pm$0.64          & 40.51$\pm$0.52          & -31.76$\pm$1.74                                                            & 33.42$\pm$0.00          & 27.62$\pm$2.36          & 23.13$\pm$1.91          & 19.72$\pm$3.02          & 25.97$\pm$2.14          & -17.75$\pm$2.48                                                            \\
F2M\cite{Shi2021OvercomingCF}                     &                      & 56.22$\pm$0.00          & 42.66$\pm$1.15          & 32.45$\pm$1.09          & 27.06$\pm$0.80          & 39.60$\pm$0.89          & -32.67$\pm$1.88                                                            & 16.79$\pm$0.00          & 11.05$\pm$1.12          & 7.30$\pm$1.90           & 4.43$\pm$1.51           & 9.89$\pm$1.34           & -33.83$\pm$1.84                                                            \\
FSLL\cite{Mazumder2021FewShotLL}                    &                      & 57.78$\pm$0.00          & 39.43$\pm$0.65          & 26.43$\pm$2.44          & 26.16$\pm$1.39          & 37.45$\pm$1.44          & -34.82$\pm$2.20                                                            & 35.27$\pm$0.00          & 25.39$\pm$0.60          & 23.62$\pm$1.52          & 22.65$\pm$1.77          & 26.73$\pm$1.20          & -16.99$\pm$1.74                                                            \\
ICARL\cite{Rebuffi2017iCaRLIC}                   &                      & 36.54$\pm$0.12          & 18.75$\pm$0.00          & 13.95$\pm$0.00          & 11.11$\pm$0.00          & 20.09$\pm$0.06          & -52.18$\pm$1.66                                                            & 25.77$\pm$0.67          & 31.29$\pm$0.00          & 29.44$\pm$0.00          & 29.06$\pm$0.00          & 28.89$\pm$0.34          & -14.83$\pm$1.31                                                            \\
Rebalance\cite{Hou2019LearningAU}                &                      & 26.09$\pm$0.00          & 18.75$\pm$0.00          & 13.95$\pm$0.00          & 11.11$\pm$0.00          & 17.48$\pm$0.00          & -54.79$\pm$1.66                                                            & 34.31$\pm$0.00          & 31.29$\pm$0.00          & 29.44$\pm$0.00          & 29.06$\pm$0.00          & 31.03$\pm$0.00          & -12.69$\pm$1.26                                                            \\
S3C\cite{Kalla2023S3CSS}                     &                      & 84.26$\pm$0.00          & 66.11$\pm$0.93          & 53.20$\pm$1.40          & 44.13$\pm$0.89          & 61.93$\pm$0.95          & -10.34$\pm$1.91                                                            & 49.86$\pm$0.00          & 34.49$\pm$0.86          & 29.74$\pm$1.20          & 28.62$\pm$1.14          & 35.68$\pm$0.93          & -8.04$\pm$1.57                                                             \\
SCN\cite{Ye2022BetterMB}                     &                      & 67.35$\pm$0.00          & 27.20$\pm$0.72          & 18.40$\pm$2.98          & 13.10$\pm$1.67          & 31.51$\pm$1.75          & -40.76$\pm$2.41                                                            & 47.39$\pm$0.00          & 8.79$\pm$0.00           & 6.62$\pm$1.46           & 3.13$\pm$0.80           & 16.48$\pm$0.83          & -27.24$\pm$1.51                                                            \\
\textcolor{orange}{ORCO\cite{ahmed2024orco}}                    &                      & 71.70$\pm$0.00          & 26.41$\pm$1.73          & 26.27$\pm$4.56          & 19.08$\pm$1.61          & 35.87$\pm$2.57          & -36.40$\pm$3.06                                                            & 36.37$\pm$0.00          & 7.86$\pm$0.59           & 4.65$\pm$1.36           & 1.70$\pm$0.84           & 12.65$\pm$0.85          & -31.07$\pm$1.50                                                            \\
\textbf{UNISA(Ours)}    &                      & \textbf{96.70$\pm$0.00} & \textbf{78.42$\pm$0.95} & \textbf{62.93$\pm$2.06} & \textbf{51.01$\pm$2.43} & \textbf{72.27$\pm$1.66} & \textbf{0.00$\pm$0.00}                                                     & \textbf{52.75$\pm$0.00} & \textbf{45.15$\pm$0.97} & \textbf{39.85$\pm$1.26} & \textbf{37.14$\pm$1.96} & \textbf{43.72$\pm$1.26} & \textbf{0.00$\pm$0.00}                                                     \\ \hline
Baseline\cite{Shi2021OvercomingCF}                & \multirow{9}{*}{3}   & 53.39$\pm$0.00          & 46.48$\pm$0.98          & 33.54$\pm$1.60          & 27.80$\pm$0.73          & 40.30$\pm$1.01          & -30.92$\pm$2.69                                                            & 33.42$\pm$0.00          & 29.85$\pm$0.58          & 23.95$\pm$2.22          & 23.01$\pm$2.67          & 27.56$\pm$1.76          & -16.46$\pm$1.91                                                            \\
F2M\cite{Shi2021OvercomingCF}                     &                      & 56.22$\pm$0.00          & 42.56$\pm$1.07          & 32.61$\pm$0.67          & 26.51$\pm$0.63          & 39.48$\pm$0.71          & -31.74$\pm$2.59                                                            & 16.79$\pm$0.00          & 9.57$\pm$2.46           & 4.95$\pm$1.28           & 5.32$\pm$1.24           & 9.16$\pm$1.52           & -34.86$\pm$1.69                                                            \\
FSLL\cite{Mazumder2021FewShotLL}                    &                      & 57.78$\pm$0.00          & 39.48$\pm$1.37          & 27.23$\pm$0.94          & 25.77$\pm$1.14          & 37.57$\pm$1.01          & -33.65$\pm$2.69                                                            & 35.27$\pm$0.00          & 26.05$\pm$1.71          & 23.88$\pm$0.37          & 23.81$\pm$0.82          & 27.25$\pm$0.97          & -16.77$\pm$1.22                                                            \\
ICARL\cite{Rebuffi2017iCaRLIC}                   &                      & 36.63$\pm$0.20          & 18.75$\pm$0.00          & 13.95$\pm$0.00          & 11.11$\pm$0.00          & 20.11$\pm$0.10          & -51.11$\pm$2.49                                                            & 26.98$\pm$1.73          & 31.29$\pm$0.00          & 29.44$\pm$0.00          & 29.06$\pm$0.00          & 29.19$\pm$0.87          & -14.83$\pm$1.14                                                            \\
Rebalance\cite{Hou2019LearningAU}               &                      & 26.09$\pm$0.00          & 18.75$\pm$0.00          & 13.95$\pm$0.00          & 11.11$\pm$0.00          & 17.48$\pm$0.00          & -53.74$\pm$2.49                                                            & 34.31$\pm$0.00          & 31.29$\pm$0.00          & 29.44$\pm$0.00          & 29.06$\pm$0.00          & 31.03$\pm$0.00          & -12.99$\pm$0.74                                                            \\
S3C\cite{Kalla2023S3CSS}                     &                      & 84.26$\pm$0.00          & 66.03$\pm$0.40          & 52.15$\pm$1.30          & 43.81$\pm$1.51          & 61.56$\pm$1.02          & -9.66$\pm$2.69                                                             & 49.86$\pm$0.00          & 35.05$\pm$0.64          & 30.50$\pm$0.94          & 28.23$\pm$1.83          & 35.91$\pm$1.08          & -8.11$\pm$1.31                                                             \\
SCN\cite{Ye2022BetterMB}                     &                      & 67.35$\pm$0.00          & 26.39$\pm$0.92          & 19.57$\pm$2.17          & 17.13$\pm$1.87          & 32.61$\pm$1.50          & -38.61$\pm$2.91                                                            & 47.39$\pm$0.00          & 8.75$\pm$0.02           & 7.21$\pm$1.30           & 4.62$\pm$1.05           & 16.99$\pm$0.84          & -27.03$\pm$1.12                                                            \\
\textcolor{orange}{ORCO\cite{ahmed2024orco}}                    &                      & 71.70$\pm$0.00          & 24.26$\pm$2.64          & 23.63$\pm$5.92          & 19.44$\pm$0.41          & 34.76$\pm$3.25          & -36.46$\pm$4.09                                                            & 36.37$\pm$0.00          & 7.86$\pm$0.59           & 4.65$\pm$1.36           & 1.70$\pm$0.84           & 12.65$\pm$0.85          & -31.37$\pm$1.13                                                            \\
\textbf{UNISA(Ours)}    &                      & \textbf{96.70$\pm$0.00} & \textbf{78.88$\pm$1.01} & \textbf{60.47$\pm$2.44} & \textbf{48.81$\pm$4.22} & \textbf{71.22$\pm$2.49} & \textbf{0.00$\pm$0.00}                                                     & \textbf{52.75$\pm$0.00} & \textbf{44.90$\pm$0.65} & \textbf{40.49$\pm$1.08} & \textbf{37.93$\pm$0.76} & \textbf{44.02$\pm$0.74} & \textbf{0.00$\pm$0.00}                                                     \\ \hline
Baseline\cite{Shi2021OvercomingCF}                & \multirow{9}{*}{2}   & 53.39$\pm$0.00          & 45.09$\pm$0.96          & 32.75$\pm$2.52          & 26.40$\pm$1.54          & 39.41$\pm$1.55          & -31.88$\pm$2.43                                                            & 33.42$\pm$0.00          & 30.40$\pm$0.42          & 24.87$\pm$2.27          & 23.43$\pm$1.55          & 28.03$\pm$1.39          & -15.61$\pm$1.55                                                            \\
F2M\cite{Shi2021OvercomingCF}                     &                      & 56.22$\pm$0.00          & 40.83$\pm$1.39          & 32.21$\pm$0.74          & 27.11$\pm$1.04          & 39.09$\pm$0.94          & -32.20$\pm$2.09                                                            & 16.79$\pm$0.00          & 9.87$\pm$1.83           & 6.46$\pm$1.15           & 5.53$\pm$1.19           & 9.66$\pm$1.23           & -33.98$\pm$1.41                                                            \\
FSLL\cite{Mazumder2021FewShotLL}                    &                      & 57.78$\pm$0.00          & 39.88$\pm$1.22          & 26.98$\pm$1.01          & 24.21$\pm$1.29          & 37.21$\pm$1.02          & -34.08$\pm$2.13                                                            & 35.27$\pm$0.00          & 25.87$\pm$2.71          & 24.89$\pm$1.04          & 23.10$\pm$1.31          & 27.28$\pm$1.59          & -16.36$\pm$1.73                                                            \\
ICARL\cite{Rebuffi2017iCaRLIC}                   &                      & 36.61$\pm$0.13          & 18.75$\pm$0.00          & 13.95$\pm$0.00          & 11.11$\pm$0.00          & 20.11$\pm$0.07          & -51.18$\pm$1.87                                                            & 24.09$\pm$5.44          & 31.29$\pm$0.00          & 29.44$\pm$0.00          & 29.06$\pm$0.00          & 28.47$\pm$2.72          & -15.17$\pm$2.80                                                            \\
Rebalance\cite{Hou2019LearningAU}               &                      & 26.09$\pm$0.00          & 18.75$\pm$0.00          & 13.95$\pm$0.00          & 11.11$\pm$0.00          & 17.48$\pm$0.00          & -53.81$\pm$1.87                                                            & 34.31$\pm$0.00          & 31.29$\pm$0.00          & 29.44$\pm$0.00          & 29.06$\pm$0.00          & 31.03$\pm$0.00          & -12.61$\pm$0.68                                                            \\
S3C\cite{Kalla2023S3CSS}                     &                      & 84.26$\pm$0.00          & 65.30$\pm$1.17          & 51.85$\pm$2.18          & 42.75$\pm$2.46          & 61.04$\pm$1.74          & -10.25$\pm$2.55                                                            & 49.86$\pm$0.00          & 34.54$\pm$0.99          & 29.61$\pm$1.20          & 27.58$\pm$1.15          & 35.4$\pm$0.97           & -8.24$\pm$1.18                                                             \\
SCN\cite{Ye2022BetterMB}                     &                      & 67.35$\pm$0.00          & 24.76$\pm$2.23          & 13.67$\pm$1.36          & 16.29$\pm$2.64          & 30.52$\pm$1.86          & -40.77$\pm$2.64                                                            & 47.48$\pm$0.00          & 8.54$\pm$0.37           & 5.26$\pm$2.51           & 3.07$\pm$1.42           & 16.09$\pm$1.45          & -27.55$\pm$1.60                                                            \\
\textcolor{orange}{ORCO\cite{ahmed2024orco}}           &                      & 71.70$\pm$0.00          & 22.42$\pm$2.46          & 20.08$\pm$2.32          & 19.20$\pm$5.55          & 33.35$\pm$3.25          & -37.94$\pm$3.75                                                            & 36.37$\pm$0.00          & 7.72$\pm$0.45           & 4.55$\pm$1.56           & 1.79$\pm$1.16           & 12.61$\pm$1.00          & -31.03$\pm$1.21                                                            \\
\textbf{UNISA(Ours)}    &                      & \textbf{96.70$\pm$0.00} & \textbf{76.84$\pm$0.91} & \textbf{61.54$\pm$1.73} & \textbf{50.08$\pm$3.19} & \textbf{71.29$\pm$1.87} & \textbf{0.00$\pm$0.00}                                                     & \textbf{52.75$\pm$0.00} & \textbf{44.32$\pm$1.05} & \textbf{39.78$\pm$0.77} & \textbf{37.72$\pm$0.36} & \textbf{43.64$\pm$0.68} & \textbf{0.00$\pm$0.00}                                                     \\ \hline
Baseline\cite{Shi2021OvercomingCF}                & \multirow{9}{*}{1}   & 53.39$\pm$0.00          & 41.69$\pm$3.05          & 31.10$\pm$1.80          & 25.56$\pm$1.34          & 37.94$\pm$1.89          & -31.92$\pm$2.52                                                            & 33.42$\pm$0.00          & 30.45$\pm$0.39          & 27.97$\pm$0.29          & 27.46$\pm$0.53          & 29.83$\pm$0.36          & -15.03$\pm$1.03                                                            \\
F2M\cite{Shi2021OvercomingCF}                     &                      & 56.22$\pm$0.00          & 40.55$\pm$1.66          & 31.21$\pm$1.02          & 24.71$\pm$1.50          & 38.17$\pm$1.23          & -31.69$\pm$2.07                                                            & 16.79$\pm$0.00          & 10.98$\pm$1.65          & 8.35$\pm$1.70           & 6.30$\pm$0.98           & 10.61$\pm$1.28          & -34.25$\pm$1.60                                                            \\
FSLL\cite{Mazumder2021FewShotLL}                    &                      & 57.78$\pm$0.00          & 38.79$\pm$1.68          & 26.51$\pm$1.91          & 22.16$\pm$1.71          & 36.31$\pm$1.53          & -33.55$\pm$2.26                                                            & 35.27$\pm$0.00          & 24.75$\pm$1.99          & 25.20$\pm$1.56          & 23.97$\pm$1.01          & 27.30$\pm$1.36          & -17.56$\pm$1.66                                                            \\
ICARL\cite{Rebuffi2017iCaRLIC}                   &                      & 36.49$\pm$0.14          & 18.75$\pm$0.00          & 13.95$\pm$0.00          & 11.11$\pm$0.00          & 20.08$\pm$0.07          & -49.78$\pm$1.66                                                            & 21.72$\pm$7.02          & 31.29$\pm$0.00          & 29.44$\pm$0.00          & 29.06$\pm$0.00          & 27.88$\pm$3.51          & -16.98$\pm$3.64                                                            \\
Rebalance\cite{Hou2019LearningAU}               &                      & 26.09$\pm$0.00          & 18.75$\pm$0.00          & 13.95$\pm$0.00          & 11.11$\pm$0.00          & 17.48$\pm$0.00          & -52.38$\pm$1.66                                                            & 34.31$\pm$0.00          & 31.29$\pm$0.00          & 29.44$\pm$0.00          & 29.06$\pm$0.00          & 31.03$\pm$0.00          & -13.83$\pm$0.96                                                            \\
S3C\cite{Kalla2023S3CSS}                     &                      & 84.26$\pm$0.00          & 63.38$\pm$2.29          & 49.35$\pm$2.04          & 39.71$\pm$1.96          & 59.18$\pm$1.82          & -10.68$\pm$2.46                                                            & 49.86$\pm$0.00          & 33.57$\pm$3.28          & 27.51$\pm$2.31          & 26.41$\pm$1.77          & 34.34$\pm$2.19          & -10.52$\pm$2.39                                                            \\
SCN\cite{Ye2022BetterMB}                     &                      & 67.35$\pm$0.00          & 24.39$\pm$1.15          & 14.48$\pm$1.21          & 10.61$\pm$1.98          & 29.21$\pm$1.29          & -40.65$\pm$2.10                                                            & 47.48$\pm$0.00          & 7.99$\pm$1.24           & 3.51$\pm$1.14           & 2.60$\pm$1.26           & 15.4$\pm$1.05           & -29.46$\pm$1.42                                                            \\
\textcolor{orange}{ORCO\cite{ahmed2024orco}}                    &                      & 71.70$\pm$0.00          & 24.67$\pm$0.99          & 19.03$\pm$1.25          & 18.32$\pm$0.77          & 33.43$\pm$0.89          & -36.43$\pm$1.88                                                            & 36.37$\pm$0.00          & 7.36$\pm$0.61           & 3.60$\pm$0.55           & 0.93$\pm$0.08           & 12.07$\pm$0.41          & -32.79$\pm$1.04                                                            \\
\textbf{UNISA(Ours)}    &                      & \textbf{96.70$\pm$0.00} & \textbf{75.06$\pm$1.56} & \textbf{59.69$\pm$1.50} & \textbf{48.00$\pm$2.52} & \textbf{69.86$\pm$1.66} & \textbf{0.00$\pm$0.00}                                                     & \textbf{52.75$\pm$0.00} & \textbf{44.80$\pm$0.70} & \textbf{41.49$\pm$1.15} & \textbf{40.38$\pm$1.37} & \textbf{44.86$\pm$0.96} & \textbf{0.00$\pm$0.00}    \\ \hline                                                              
\end{tabular}
\caption{\textcolor{blue}{Classification accuracy on split EuroSAT and HyRank dataset with 1-5 shot setting
averaged across 5 times run.}}
\label{tab:eurosat_and_hyrank}
\end{table*}

\begin{table*}[]
\centering
\scriptsize
\begin{tabular}{ccccccccccc}
\hline
{}  & \multicolumn{5}{c}{{Configuration}}     & \multicolumn{4}{c}{{Session}}   & {} \\ \cline{2-10}
\multirow{-2}{*}{{Code}} & {FM} & {WM ($\mathcal{L}_{cpr}$)} & {$\mathcal{L}_{psl}$} & {$\mathcal{L}_{psa}$} & {$\mathcal{L}_{ball}$} & {1}              & {2}              & {3}              & {4}              & \multirow{-2}{*}{{Avg}} \\ \hline
A                     &     & $\checkmark$   & $\checkmark$     & $\checkmark$     & $\checkmark$     & 65.00$\pm$0.00          & 55.80$\pm$1.35          & 48.83$\pm$1.19          & 40.95$\pm$1.05          & 52.65$\pm$1.04          \\
B                     & $\checkmark$   &     & $\checkmark$     & $\checkmark$     & $\checkmark$     & 62.92$\pm$0.00          & 56.27$\pm$1.98          & 50.22$\pm$1.48          & 43.33$\pm$0.66          & 53.19$\pm$1.28          \\
C                     & $\checkmark$   & $\checkmark$   &       & $\checkmark$     & $\checkmark$     & 65.83$\pm$0.00          & 55.20$\pm$0.86          & 46.17$\pm$0.97          & 38.43$\pm$1.01          & 51.41$\pm$0.82          \\
D                     & $\checkmark$   & $\checkmark$   & $\checkmark$     &       & $\checkmark$     & 51.67$\pm$0.00          & 45.80$\pm$1.74          & 41.33$\pm$1.56          & 35.10$\pm$2.62          & 43.48$\pm$1.76          \\
E                     & $\checkmark$   & $\checkmark$   & $\checkmark$     & $\checkmark$     &       & 66.67$\pm$0.00          & 56.93$\pm$0.34          & 47.00$\pm$1.89          & 40.05$\pm$1.45          & 52.66$\pm$1.20          \\
UNISA                 & $\checkmark$   & $\checkmark$   & $\checkmark$     & $\checkmark$     & $\checkmark$     & \textbf{67.50$\pm$0.00} & \textbf{58.27$\pm$1.51} & \textbf{51.67$\pm$0.77} & \textbf{43.76$\pm$1.50} & \textbf{55.30$\pm$1.13} \\ \hline
\end{tabular}
\caption{\textcolor{blue}{Ablation study of the proposed method with various configurations on 5-shot UCMerced dataset, FM and WM stand for Flat Minima and Wide Minima respectively.}}
\label{tab:ablation}
\end{table*}

\begin{table}[]
\centering
\scriptsize
\begin{tabular}{c|ccc|ccc}
\hline
Code        & $\lambda_1$           & $\lambda_2$           & $\lambda_3$           & $\lambda_4$           & $\lambda_5$           & Avg. Accuracy               \\ \hline
\textbf{U1} & \textbf{1.0} & \textbf{1.0} & \textbf{1.0} & \textbf{0.1} & \textbf{1.0} & \textbf{*55.30+-1.13} \\
U2          & 1.0          & 1.0          & 0.1          & 0.1          & 1.0          & 54.35+-1.47           \\
U3          & 1.0          & 0.1          & 1.0          & 0.1          & 1.0          & 54.35+-1.47           \\
U4          & 1.0          & 0.1          & 0.1          & 0.1          & 1.0          & 54.35+-1.47           \\
U5          & 0.1          & 1.0          & 1.0          & 0.1          & 1.0          & 49.16+-1.02           \\
U6          & 0.1          & 1.0          & 0.1          & 0.1          & 1.0          & 49.16+-1.02           \\
U7          & 0.1          & 0.1          & 1.0          & 0.1          & 1.0          & 49.16+-1.02           \\
U8          & 0.1          & 0.1          & 0.1          & 0.1          & 1.0          & 49.16+-1.02           \\ \hline
U9          & 1.0          & 1.0          & 1.0          & 0.1          & 0.1          & 55.31+-1.14           \\
U10         & 1.0          & 1.0          & 1.0          & 0.01         & 1.0          & 55.31+-1.14           \\
U11         & 1.0          & 1.0          & 1.0          & 0.01         & 0.1          & 55.31+-1.14          \\ \hline
\end{tabular}
\caption{\textcolor{blue}{Extended ablation study of the proposed method with various loss coefficients i.e. $\lambda_i, i \in [1-5]$ on 5-shot UCMerced dataset, * denotes the result of our default configuration. }}
\label{tab:balance}
\end{table}

\begin{table}[]
\centering
\setlength{\tabcolsep}{0.093em}
\scriptsize
\begin{tabular}{lcccccc}
\hline
\multirow{2}{*}{Method} & \multirow{2}{*}{Noise} & \multicolumn{4}{c}{Session}                                                               & \multirow{2}{*}{Avg} \\ \cline{3-6}
                        &                        & 1                    & 2                    & 3                    & 4                    &                      \\ \hline
{UNISA(Ours)}          & -                      & \textbf{67.50$\pm$0.00} & \textbf{58.27$\pm$1.51} & \textbf{51.67$\pm$0.77} & \textbf{43.76$\pm$1.50} & \textbf{55.30$\pm$1.13} \\
UNISA(Ours)                   & Brightness             & 67.50$\pm$0.00          & 58.27$\pm$1.51          & 51.67$\pm$0.77          & 43.76$\pm$1.50          & 55.30$\pm$1.13          \\
UNISA(Ours)                   & Contrass               & 67.50$\pm$0.00          & 58.27$\pm$1.51          & 51.67$\pm$0.77          & 43.76$\pm$1.50          & 55.30$\pm$1.13          \\
UNISA(Ours)                   & Hue                    & 67.50$\pm$0.00          & 58.27$\pm$1.51          & 51.67$\pm$0.77          & 43.76$\pm$1.50          & 55.30$\pm$1.13          \\
UNISA(Ours)                   & Saturation             & 67.50$\pm$0.00          & 58.27$\pm$1.51          & 51.67$\pm$0.77          & 43.76$\pm$1.50          & 55.30$\pm$1.13          \\
UNISA(Ours)                   & Blur                   & 67.50$\pm$0.00          & 58.27$\pm$1.51          & 51.67$\pm$0.77          & 43.76$\pm$1.50          & 55.30$\pm$1.13          \\
UNISA(Ours)                   & Gaussian               & 66.25$\pm$0.00          & 57.67$\pm$0.92          & 49.72$\pm$1.85          & 42.86$\pm$1.11          & 54.13$\pm$1.17          \\ \hline
Baseline\cite{Shi2021OvercomingCF}                & Gaussian               & 19.58$\pm$0.00          & 17.33$\pm$0.87          & 14.22$\pm$1.44          & 11.71$\pm$0.95          & 15.71$\pm$0.97          \\
F2M\cite{Shi2021OvercomingCF}                     & Gaussian               & 25.42$\pm$0.00          & 16.27$\pm$0.39          & 13.39$\pm$1.27          & 10.81$\pm$1.13          & 16.47$\pm$0.87          \\
FSLL\cite{Mazumder2021FewShotLL}                    & Gaussian               & 24.17$\pm$0.00          & 11.73$\pm$1.29          & 9.39$\pm$1.26           & 7.90$\pm$1.01           & 13.30$\pm$1.03          \\
ICARL\cite{Rebuffi2017iCaRLIC}                   & Gaussian               & 26.00$\pm$2.55          & 20.40$\pm$0.50          & 17.28$\pm$1.39          & 14.43$\pm$0.86          & 19.53$\pm$1.53          \\
Rebalance\cite{Hou2019LearningAU}               & Gaussian               & 13.33$\pm$0.40          & 10.80$\pm$0.40          & 9.00$\pm$0.12           & 7.57$\pm$0.16           & 10.18$\pm$0.30          \\
S3C\cite{Kalla2023S3CSS}                     & Gaussian               & 53.75$\pm$0.00          & 41.40$\pm$1.79          & 33.11$\pm$0.56          & 30.14$\pm$1.45          & 39.60$\pm$1.19          \\
SCN\cite{Ye2022BetterMB}                     & Gaussian               & 21.67$\pm$0.00          & 10.67$\pm$1.57          & 8.61$\pm$1.90           & 8.76$\pm$1.11           & 12.43$\pm$1.35          \\
\textcolor{orange}{ORCO\cite{ahmed2024orco}}                    & Gaussian               & 35.83$\pm$0.00          & 10.67$\pm$2.49          & 14.89$\pm$3.32          & 15.62$\pm$2.74          & 19.25$\pm$2.49         \\ \hline
\end{tabular}
\caption{\textcolor{blue}{Robustness analysis of the proposed method against various noises on 5-shot UCMerced dataset.}}
\label{tab:robustness_againts_noise}
\end{table}

\begin{table}[]
\setlength{\tabcolsep}{0.25em}
\scriptsize
\centering
\begin{tabular}{l|c|ccccc}
\hline
\multicolumn{2}{c}{Hyper Params.} & \multicolumn{4}{c}{Session}                           & \multirow{2}{*}{Avg} \\ \cline{1-6}
Min.LR             & BS            & 1           & 2           & 3           & 4           &                      \\ \hline
0.001             & 128           & 67.50$\pm$0.00 & 58.27$\pm$1.51 & 51.67$\pm$0.77 & 43.76$\pm$1.50 & 55.30$\pm$1.13           \\
0.0001            & 128           & 67.50$\pm$0.00 & 58.27$\pm$1.51 & 51.67$\pm$0.77 & 43.76$\pm$1.50 & 55.30$\pm$1.13           \\
0.00001           & 128           & 67.50$\pm$0.00 & 58.27$\pm$1.51 & 51.67$\pm$0.77 & 43.76$\pm$1.50 & 55.30$\pm$1.13           \\
0.0               & 128           & 67.50$\pm$0.00 & 58.27$\pm$1.51 & 51.67$\pm$0.77 & 43.76$\pm$1.50 & 55.30$\pm$1.13           \\
0.0               & 64            & 67.50$\pm$0.00 & 58.27$\pm$1.51 & 51.67$\pm$0.77 & 43.76$\pm$1.50 & 55.30$\pm$1.13           \\
0.0               & 32            & 67.50$\pm$0.00 & 58.27$\pm$1.51 & 51.67$\pm$0.77 & 43.76$\pm$1.50 & 55.30$\pm$1.13           \\
0.0               & 16            & 67.50$\pm$0.00 & 58.27$\pm$1.51 & 51.67$\pm$0.77 & 43.76$\pm$1.50 & 55.30$\pm$1.13           \\
0.0               & 8             & 67.50$\pm$0.00 & 57.20$\pm$1.74 & 49.61$\pm$1.79 & 43.38$\pm$0.96 & 54.42$\pm$1.00          \\
0.0               & 4             & 67.50$\pm$0.00 & 55.40$\pm$2.00 & 47.56$\pm$2.16 & 40.10$\pm$1.47 & 52.64$\pm$1.64 \\ \hline         
\end{tabular}
\caption{\textcolor{blue}{Robustness analysis of the proposed method in various hyperparameters settings on 5-shot UCMerced dataset. Min.LR and BS denote minimum learning rate and batch-size respectively.}
\label{tab:robustness_againts_hyperparams}}
\end{table}

\begin{table}[]
\centering
\setlength{\tabcolsep}{0.3em}
\scriptsize
\centering
\begin{tabular}{lccccc}
\hline
{}                            & \multicolumn{4}{c}{{Session}}  & {}                          \\ \cline{2-5}
\multirow{-2}{*}{{Bound (b)}} & {1}     & {2}     & {3}     & {4}     & \multirow{-2}{*}{{Average}} \\ \hline
0.001 & 69.58$\pm$0.00 & 57.73$\pm$1.85 & 51.11$\pm$1.35 & 43.48$\pm$1.23 & 55.48$\pm$1.30 \\
0.005 & 69.58$\pm$0.00 & 58.00$\pm$1.81 & 51.11$\pm$0.92 & 43.86$\pm$1.36 & 55.64$\pm$1.22 \\
0.01  & 67.50$\pm$0.00 & 58.27$\pm$1.51 & 51.67$\pm$0.77 & 43.76$\pm$1.50 & 55.30$\pm$1.13 \\
0.05  & 69.58$\pm$0.00 & 58.00$\pm$1.81 & 51.11$\pm$0.92 & 43.86$\pm$1.36 & 55.64$\pm$1.22 \\
0.1   & 68.33$\pm$0.00 & 58.07$\pm$1.20 & 50.61$\pm$0.42 & 42.43$\pm$0.52 & 54.86$\pm$0.69 \\ \hline
\end{tabular}
\caption{\textcolor{blue}{Sensitivity analysis of of the proposed  with various bound (b) value on 5-shot UCMerced dataset}}
\label{tab:bound_analysis}
\end{table}

\begin{table}[]
\centering
\setlength{\tabcolsep}{0.3em}
\scriptsize
\centering
\begin{tabular}{cccccc}
\hline
{}  & \multicolumn{4}{c}{{Session}}                                                                        & {}                          \\ \cline{2-5}
\multirow{-2}{*}{{\begin{tabular}[c]{@{}c@{}}Perturbation \\ loop (M)\end{tabular}}} & {1}     & {2}     & {3}     & {4}     & \multirow{-2}{*}{{Average}} \\ \hline
1 & 64.58$\pm$0.00 & 55.87$\pm$1.43 & 48.94$\pm$1.09 & 40.90$\pm$0.95 & 52.57$\pm$1.02 \\
2 & 67.50$\pm$0.00 & 58.27$\pm$1.51 & 51.67$\pm$0.77 & 43.76$\pm$1.50 & 55.30$\pm$1.13 \\
3 & 65.42$\pm$0.00 & 57.07$\pm$1.40 & 49.33$\pm$1.75 & 41.57$\pm$1.63 & 53.35$\pm$1.39 \\
4 & 67.50$\pm$0.00 & 60.27$\pm$1.29 & 53.39$\pm$0.81 & 46.00$\pm$1.34 & 56.79$\pm$1.01 \\
5 & 72.50$\pm$0.00 & 58.93$\pm$2.04 & 50.22$\pm$1.01 & 43.43$\pm$0.41 & 56.27$\pm$1.16 \\
6 & 75.83$\pm$0.00 & 63.20$\pm$1.59 & 54.28$\pm$1.22 & 45.57$\pm$0.99 & 59.72$\pm$1.12 \\
7 & 79.17$\pm$0.00 & 66.07$\pm$1.11 & 56.78$\pm$1.33 & 47.86$\pm$0.48 & 62.47$\pm$0.90  \\ \hline          \end{tabular}     
\caption{\textcolor{blue}{Impact of number of perturbation loop (M) to the proposed method performance  on 5-shot UCMerced dataset}}
\label{tab:M_analysis}
\end{table}

\begin{figure*}
\includegraphics[width=1.0\textwidth]{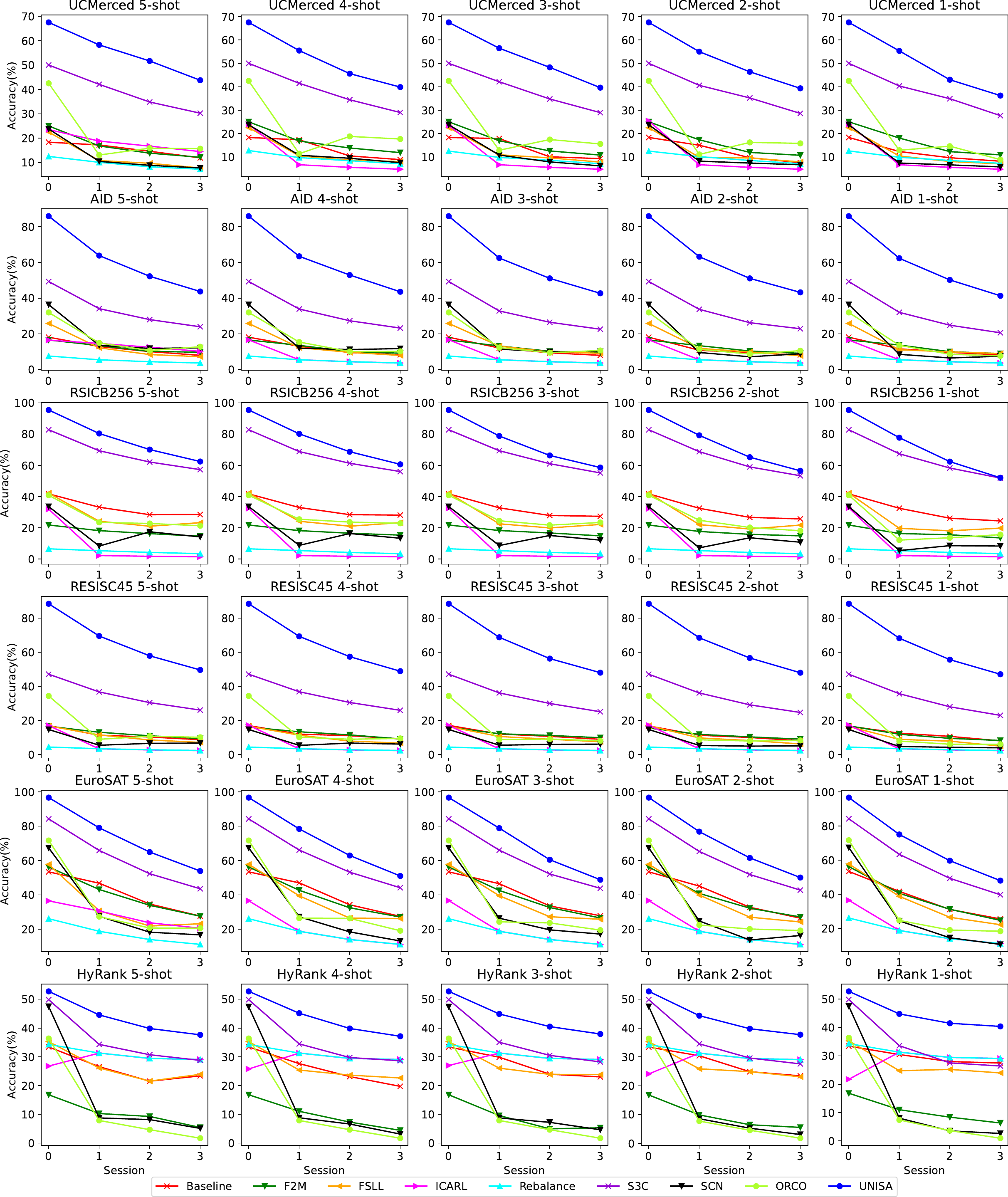}
\caption{\textcolor{blue}{Visualization of Performances of Consolidated Algorithms across 6 dataset in 1-5 shot settings}}
\label{Fig:4dataset_plot}
\end{figure*}

\begin{table*}[]
\centering
\scriptsize
\setlength{\tabcolsep}{0.19em}
\begin{tabular}{l|cc|cc|cc|cc|cc|cc}
\hline
\multirow{3}{*}{Method} & \multicolumn{10}{c}{Training time per epoch (s)}                                                                                                       &            &               \\ \cline{2-13}
                        & \multicolumn{2}{c}{UCMerced} & \multicolumn{2}{c}{AID}     & \multicolumn{2}{c}{RSICB256} & \multicolumn{2}{c}{RESISC45} & \multicolumn{2}{c}{EuroSAT} & \multicolumn{2}{c}{HyRank} \\ \cline{1-13}
                        & Base task   & Few-shot task  & Base task   & Few-shot task & Base task    & Few-shot task & Base task    & Few-shot task & Base task   & Few-shot task & Base task  & Few-shot task \\ \hline
Baseline\cite{Shi2021OvercomingCF}                & 0.54$\pm$0.01  & 0.04$\pm$0.00     & 1.33$\pm$0.02  & 0.09$\pm$0.02    & 0.73$\pm$0.02   & 0.15$\pm$0.01    & 0.96$\pm$0.09   & 0.11$\pm$0.02    & 0.65$\pm$0.01  & 0.15$\pm$0.01    & 0.62$\pm$0.01 & 0.10$\pm$0.00    \\
F2M\cite{Shi2021OvercomingCF}                     & 0.64$\pm$0.01  & 0.73$\pm$0.05     & 1.38$\pm$0.06  & 1.11$\pm$0.08    & 1.54$\pm$0.01   & 0.57$\pm$0.03    & 1.13$\pm$0.06   & 0.41$\pm$0.02    & 0.84$\pm$0.01  & 0.68$\pm$0.05    & 0.65$\pm$0.00 & 0.78$\pm$0.02    \\
FSLL\cite{Mazumder2021FewShotLL}                    & 0.54$\pm$0.01  & 0.68$\pm$0.07     & 1.33$\pm$0.02  & 1.36$\pm$0.04    & 0.73$\pm$0.02   & 1.02$\pm$0.05    & 0.96$\pm$0.09   & 0.36$\pm$0.05    & 0.65$\pm$0.01  & 0.7$\pm$0.08     & 0.62$\pm$0.01 & 0.67$\pm$0.02    \\
ICARL\cite{Rebuffi2017iCaRLIC}                   & 0.59$\pm$0.02  & 0.30$\pm$0.01     & 1.27$\pm$0.04  & 0.62$\pm$0.10    & 0.71$\pm$0.03   & 1.64$\pm$0.16    & 0.67$\pm$0.07   & 1.77$\pm$0.15    & 0.79$\pm$0.10  & 1.37$\pm$0.06    & 0.60$\pm$0.00 & 1.35$\pm$0.04    \\
Rebalance\cite{Hou2019LearningAU}               & 0.58$\pm$0.02  & 0.34$\pm$0.00     & 1.15$\pm$0.01  & 1.67$\pm$0.42    & 0.81$\pm$0.04   & 1.64$\pm$0.16    & 0.65$\pm$0.10   & 1.68$\pm$0.13    & 0.63$\pm$0.01  & 1.26$\pm$0.02    & 0.79$\pm$0.00 & 1.27$\pm$0.01    \\
S3C\cite{Kalla2023S3CSS}                     & 1.18$\pm$0.05  & 0.62$\pm$0.01     & 3.14$\pm$0.04  & 0.95$\pm$0.06    & 2.11$\pm$0.35   & 0.84$\pm$0.08    & 2.90$\pm$0.05   & 1.28$\pm$0.02    & 2.41$\pm$0.72  & 0.86$\pm$0.02    & 1.51$\pm$0.09 & 0.76$\pm$0.02    \\
SCN\cite{Ye2022BetterMB}                     & 0.84$\pm$0.01  & 2.40$\pm$0.17     & 6.53$\pm$0.20  & 7.05$\pm$0.94    & 3.60$\pm$0.92   & 7.10$\pm$0.54    & 4.03$\pm$0.00   & 8.20$\pm$0.30    & 2.17$\pm$0.56  & 5.97$\pm$0.89    & 1.64$\pm$0.14 & 7.63$\pm$0.21    \\
\textcolor{orange}{ORCO\cite{ahmed2024orco}}                    & 2.20$\pm$0.13  & 1.15$\pm$0.01     & 4.31$\pm$0.22  & 1.04$\pm$0.03    & 2.59$\pm$0.14   & 0.83$\pm$0.02    & 3.62$\pm$0.75   & 1.38$\pm$0.06    & 2.28$\pm$0.22  & 1.18$\pm$0.11    & 2.71$\pm$0.36 & 1.25$\pm$0.02    \\
UNISA(Ours)                   & 2.60$\pm$0.55  & 1.88$\pm$0.18     & 11.67$\pm$0.58 & 3.52$\pm$0.56    & 20.67$\pm$0.58  & 2.34$\pm$0.50    & 29.33$\pm$0.58  & 1.67$\pm$0.58    & 17.33$\pm$0.58 & 1.58$\pm$0.37    & 4.33$\pm$0.58 & 2.11$\pm$0.36   \\ \hline
\end{tabular}
\caption{\textcolor{blue}{Running time of the consolidated algorithms}}
\label{tab:running_time}
\end{table*}

\section{Experiments}
This section presents numerical validations of UNISA, including experiments on remote sensing problems, comparisons against prior arts, ablation studies to demonstrate the advantage of each learning module and sensitivity analysis. Source codes of UNISA are shared publicly in \url{https://github.com/anwarmaxsum/UNISA}. 

\subsection{Experimental Setting}
All consolidated algorithms are benchmarked in six RS datasets: the UC-Merced \cite{Yang2010BagofvisualwordsAS}, the AID \cite{Xia2016AIDAB}, the RSICB-256 \cite{Li2017RSICBAL}, the RESISC45 \cite{Cheng2017resisc45}, the EuroSAT \cite{helber2019eurosat}, the HyRANK \cite{Zhang2021TopologicalSA} under 1-shot, 2-shot, 3-shot, 4-shot and 5-shot configurations. 

\noindent\textbf{Dataset}: the UC Merced dataset comprises 21 classes where each class carries 100 images. It is split into 4 tasks where the base task possesses 12 classes while each few shot task includes 3 classes. The spatial resolution of this dataset is 0.3 m. Visualization of the UC Merced problem is offered in Fig. \ref{Fig:ucmerceddataset}. The AID dataset presents 30 categories where each category has around ~300 images with the spatial resolution between 0.5 m and 8 m. This dataset is divided into 4 tasks where base task encompasses 15 classes while each few-shot task consists of 5 classes. The dataset, RSICB256, has 35 classes where each class comprises 800 images with spatial resolution between 0.22 m and 3 m. This dataset is set to 4 tasks where the base task consists of 20 classes and each few-shot task carries 5 classes. RESISC45 dataset carries 45 classes where each class contains 700 images with spatial resolution between 0.2 m and more than 30m/px. This dataset is set to 4 tasks where the base task consists of 24 classes and each few-shot task carries 7 classes.
The EuroSAT dataset portrays the land-use classification problem of the Sentinel 2 satellite. This dataset comprises 10 classes where the base task covers 4 classes while each few-shot tasks carries 2 classes. The HyRANK dataset is a hyperspectral dataset developed by the international society for photogrammetry and remote sensing scientific initiatives \cite{Zhang2021TopologicalSA}. The dataset is collected by the Hyperion sensor having 176 spectral bands. It consists of two scenes, Dioni and Loukia, where we focus only on the Loukia scene here having $249\times945$ pixels. This dataset comprises 12 classes where the first 6 classes are reserved for the base class while the few-shot tasks carry 2 classes totalling 3 few-shot tasks.\textcolor{orange}{We only apply a maximum $10\%$ (25 samples) label proportion of the base task while the few-shot tasks exploit the full label proportions due to the data scarcity problems. 1-5 shots cases are attempted in our experiments. Note that labeled samples are only used for inferences in UNISA while the training process itself is free of labeled samples.} 

\noindent\textbf{Baseline}: UNISA is compared against iCaRL \cite{Rebuffi2017iCaRLIC}, Rebalance \cite{Hou2019LearningAU}, FSLL \cite{Mazumder2021FewShotLL}, F2M \cite{Shi2021OvercomingCF}, S3C \cite{Kalla2023S3CSS}, and \textcolor{orange}{ORCO \cite{ahmed2024orco}} representing prior arts of the few-shot continual learning algorithms. In addition, baseline is added and reflects the case where a network is frozen while leaving only the prototype construction in the few-shot learning phase. we also compare our approach with state-of-the-art continual RS learning, namely asymmetric collaborative network (SCN) \cite{Ye2022BetterMB}. Since these algorithms are fully supervised in nature, they are only trained with $10\%$ labelled samples of the base task. To the best of our knowledge, there do not exist any algorithms designed specifically for unsupervised few-shot continual learning problems. All consolidated algorithms are executed under the same computational environments, single NVDIA A100 GPU with 40 GB RAM, using their official implementations.  \textcolor{orange}{Note that the competitor methods are trained in a supervised way utilizing all available labels both in the base task and the few shot tasks, while our proposed method (UNISA) is trained in an unsupervised way using unlabeled samples in both tasks. UNISA only utilizes labels for inference (testing).}

\noindent\textbf{Hyperparameters}: we follow the same hyperparameters setting from F2M \cite{Shi2021OvercomingCF} for competitor methods and UNISA. UNISA also follows the same hyperparameters as \cite{Huang2021LearningRF}. \textbf{Base Task}: F2M is run with 1000 epoch, 0.1 initial learning rate, 0.01 bound value, and 2 random times. Baseline and FSLL are run with 240 epoch and 0.1 initial learning rate. ICARL and Rebalance are run with 130 epoch and 0.1 initial learning rate. UNISA is run with 1000 epoch,  0.05 initial learning rate, 0.01 bound value, and 2 random times. \textbf{Few Shot Task}: F2M is run with 6 epoch, and 0.02 initial learning rate. Baseline model only generates prototypes (not updated) in few-shot task. FSLL is run with 16 epoch, and 0.001 initial learning rate. ICARL and Rebalance are run with 90 epoch and 0.01 initial learning rate. UNISA is run with maximum 5 epochs and predicted learning rate by cosine annealing. All methods use 0.9 momentum, 5e-4 weight decay, 128 bacth size, SGD optimizer, learning rate scheduler, and ResNet18 as base network. These hyper-parameters are found using the grid search to ensure fair comparisons. $\lambda_i \text{ for } i \in [1,5]$ is set to 1.0 except $\lambda_4$ that is set to 0.1. The values for the coefficient are chosen so that each term in equation (9) is in the range [0.0, 1.0].

\subsection{Numerical Results}
Table \ref{tab:ucmerced_and_aid} reports consolidated numerical results of the UC merced problem. UNISA outperforms its counterparts with large margins in 1-5 shots settings, i.e., over $~30\%$. This applies for both per task accuracy and average accuracy. The same pattern is observed in the AID dataset shown in Table \ref{tab:ucmerced_and_aid}. UNISA beats other algorithms with over $~50\%$ gaps for all cases. This finding is consistent also for the RSICB256 problem, RESISC45, and the EuroSAT problem as reported in Table \ref{tab:rsicb256_and_resisc45} and \ref{tab:eurosat_and_hyrank} where UNISA delivers the most encouraging performances with large gaps, over $30\%$ differences. In addition, UNISA performs well in the hyperspectral dataset, the HyRank dataset, with over 8\% margins compared to its competitors as reported in Table \ref{tab:eurosat_and_hyrank}. Large margins confirm the challenging characteristics of the unsupervised few-shot continual learning which cannot be handled by existing few-shot continual learning algorithms calling for fully labeled samples. To the best of our knowledge, there does not exist any direct competitors of UNISA specifically devised for unsupervised few-shot continual learning problems. In addition, RS images are also difficult to deal with because of their spatio-temporal natures. It is seen by low base task accuracy of consolidated algorithms except UNISA. On the other hand, UNISA copes with the RS data and low labels equally well where its base task accuracy is significantly better than prior arts. Fig. \ref{Fig:4dataset_plot} visualizes the trace of classification accuracy per task across all four datasets. It is seen that UNISA is superior to other seven algorithms with large margins. Its performances are consistent across all tasks and the accuracy of the base task is significantly higher than other algorithms. This fact indicates the success of UNISA in handling low label situations and spatio-temporal natures of RS images. 

\subsection{Ablation Study}
This subsection analyzes UNISA performances under five different configurations where numerical results are reported in Table \ref{tab:ablation}. Our ablation study is run under the UC merced problem.  Configuration A presents the absence of flat minima concepts, while configuration B portrays UNISA's performances without the wide minima concept. Configuration C and D respectively stand for the absence of prototype scattering loss and the positive alignment loss. Last but not least, configuration E offers the case of without the ball generator loss. It is confirmed that each learning configuration plays key roles to UNISA's performances where the absence of one loss function leads to degraded performances. The most substantial deterioration is observed in configuration D without the positive alignment loss where the performance is compromised by over $12\%$. This finding demonstrates the presence of class collision issue. The absence of prototype scattering loss in configuration C also leads to dramatic performance losses. This result signifies lacks of uniformity term bringing down generalization performances. Configuration A and B also exhibit lower performances than UNISA attributed to the catastrophic forgetting problem. Configuration E is worse by about $3\%$ than UNISA.

\subsection{\textcolor{blue}{The Balance of Loss Coefficients}}
\textcolor{blue}{We extend our ablation study to investigate the impact of loss coefficient $\lambda_i, i \in [1,5]$ on the average performance of UNISA. Table \ref{tab:balance} shows the average performance of UNISA in variations of $\lambda_i$. The upper part of the table (U1-U8) shows the impact of $\lambda_1$, $\lambda_2$, and $\lambda_3$ that is applied since the base task, while the lower part of the table (U9-U11) shows the impact of $\lambda_4$ and $\lambda_5$ that applied in few shot tasks only. The table shows that the lower value of $\lambda_1$ costs the highest performance drop i.e. about $5\%$. While the lower value of $\lambda_2$ or $\lambda_3$ decreases the average performance by about $1\%$. Please note that the zero value of $\lambda_1$ and $\lambda_2$ (the absence of $\mathcal{L}_{psa}$ and $\mathcal{L}_{cpr}$) drop the average accuracy by $12\%$ and $2\%$  respectively. It means that the lower coefficients still achieve a higher performance than zero. The table shows that the proposed method maintains its performance in a lower value of $\lambda_4$ or $\lambda_5$.}

\subsection{\textcolor{blue}{Robustness Analysis}}
\textcolor{green}{This subsection discusses the robustness analysis of the proposed method against noise attacks on testing data and different hyperparameter settings. Table \ref{tab:robustness_againts_noise} shows that the proposed method consistently maintains its performance against brightness, contrast, hue, saturation, blur, and Gaussian \cite{fu2023progressive} noises. The Gaussian noise attack reduces its performance by about $1\%$ on average, while the other noise attacks don't impact its performance. In addition, the table shows that the proposed method consistently achieves the highest performance with a significant gap ($\geq 14\%$) compared to the competitor methods under the same noise attack i.e. Gaussian noise.} \textcolor{blue}{Table \ref{tab:robustness_againts_hyperparams} shows the robustness of our proposed method under different hyperparameter settings i.e. minimum learning rate and batch size. Note that our method utilizes a learning rate predictor by cosine annealing that takes the minimum learning rate as its lower bound. The table shows that the performance of our proposed method decreases only on the smallest batch-size settings i.e. 4 and 8.}

\subsection{Sensitivity Analysis}
This subsection offers our sensitivity analysis done with the UC merced problem. Specifically, different flat minima bounds $b$ are attempted as well as the number of perturbation loops $M$. Table \ref{tab:bound_analysis} and \ref{tab:M_analysis} report our numerical results. Different flat minima bounds do not result in significant differences in performances. Nevertheless, a too-large bound causes performance deterioration because of inaccurate approximations of flat-wide minima regions contributing to the catastrophic forgetting problem. An interesting observation is found in respect to the perturbation loop $M$ where the least performance is attained with $M=1$ because there does not exist any flat-wide minima estimations. UNISA's performances constantly increase with the increase of the perturbation loop $M$, i.e., the highest performance is attained with $M=7$ since this step result in fine-grained approximations of flat-wide minima regions. Note that extra perturbation loop $M$ incurs additional computational burdens.  

\subsection{\textcolor{green}{Complexity Analysis and Execution Times}}
{\textcolor{green}{Please see our supplemental document for a detailed complexity analysis. Our complexity analysis shows that our method has $O(N)$ complexity, where $N$ is the size of the dataset accumulated from the first task until the last task.}   Table \ref{tab:running_time} reports the training time per epoch for all consolidated algorithms. UNISA imposes higher execution times than other consolidated algorithms in the base task. The reason for this is found in the perturbation mechanism of UNISA where $M$ perturbation loops are required to discover flat-wide minima regions. Note that other algorithms utilize $10\%$ of base samples due to the absence of true class labels, thus expediting their running times. On the other side, the execution times of few-shot tasks of UNISA are comparable to other algorithms and always lower than SCN. 

\section{Limitations}
UNISA has demonstrated encouraging performances compared to prior arts while addressing an unexplored problem, unsupervised few-shot continual learning (UFSCL). Nonetheless, there exist some rooms for further studies.
\begin{itemize}
    \item Although the UFSCL problem is highly challenging and concerns on the problem of label scarcities in the FSCL, all tasks are drawn from the same domain. This opens a future research issue to address the problem of cross-domain FSCL involving several domains. 
    \item Although UNISA is memory-free and unlike \cite{Shi2021OvercomingCF}, it relies on the data augmentation procedure to address a sequence of few-shot learning tasks. Such mechanism simplifies the FSCL problem to some extents. Another research issue is how to devise an augmentation-free FSCL approach. Note that \cite{Tao2020FewShotCL,Shi2021OvercomingCF,Chen2021IncrementalFL,Mazumder2021FewShotLL} usually bias toward the base task notwithstanding that they do not apply any data augmentation procedures. That is, they are only capable of classifying the base classes while performing poorly for few-shot classes. 
\end{itemize}

\section{Conclusion}
This paper proposes an extension of few-shot continual learning, namely unsupervised few-shot continual learning where no labels are exploited for inducing the training process including its solution, unsupervised flat-wide learning approach (UNISA). UNISA is built upon the flat-wide learning approach to address the CF problem while the concept of prototype scattering and positive sampling is implemented as a representation learning strategy. The efficacy of UNISA has been validated in the remote sensing image scene classification problems where it produces the most encouraging numerical results over its prior arts. Our future work is devoted to address the data privacy problem in continual learning.

\ifCLASSOPTIONcaptionsoff
  \newpage
\fi



%\bibliographystyle{IEEEtran}
\bibliographystyle{IEEEtran}
\bibliography{references}



%\vfill


\end{document}

\title{Supplemental Document for Unsupervised Few-Shot Continual Learning for Remote Sensing Image Scene Classification}

\author{Muhammad Anwar~Ma'sum,
        Mahardhika~Pratama,~\IEEEmembership{Senior Member,~IEEE}
        , Ramasamy~Savitha,~\IEEEmembership{Senior Member,~IEEE}, Lin~Liu, Habibullah, Ryszard~Kowalczyk% <-this % stops a space
\thanks{M. A. Ma'sum, M. Pratama, L. Liu, Habibullah, R. Kowalcyzk are with STEM, University of South Australia, Adelaide, South Australia, Australia. R. Savitha is with I2R, A*Star, Singapore. R. Kowalczyk is also with System Research Institute, Polish Academy of Sciences.}% <-this % stops a space
\thanks{M. A. Ma'sum and M. Pratama share equal contributions}% <-this % stops a space
\thanks{Manuscript received April 19, 2005; revised August 26, 2015.}}


\markboth{Journal of \LaTeX\ Class Files,~Vol.~14, No.~8, August~2015}%
{Shell \MakeLowercase{\textit{et al.}}: Bare Demo of IEEEtran.cls for IEEE Journals}



\maketitle

\begin{abstract}
This supplementary document presents the learning policy of UNISA that is presented  in section \ref{unisa_alg} and compelxity analysis presented in section \ref{unisa_comp}.
\end{abstract}

\begin{IEEEkeywords}
Few-shot Learning, Few-Shot Continual Learning, Continual Learning, Unsupervised Learning.
\end{IEEEkeywords}

% \appendix

\section{UNISA Algorithm}\label{unisa_alg}
The learning policy of UNISA both for base task and few shot task is presented in algorithm \ref{algorithm1}.

\begin{algorithm*}
\caption{Learning Policy of UNISA}\label{algorithm1}
\begin{algorithmic}[1]
\State \textbf{Input:} The flat-wide region bound $b$, randomly initialized parameters $\Theta=\{\Phi,\Phi', \Psi\}$, projection module parameter $w_X$, hyperparameetrs $\lambda_{1},\lambda_{2},\lambda_{3},\lambda_{4}$, number of epochs $E$, number of batches on each task $O$, number of perturbation $M$.  number of synthetic samples $S$.
\State \textbf{Output:}  Updated Network Parameters $\Theta=\{\Phi, \Phi', \Psi\}$, Updated projection module $w_X$
\item[\hspace{5 mm} //Base Learning Phase $k=1$]
\For{$e=1:E$}
    \State $PseudoLabel=KMeans\_Clustering()$
    \For{$batch=1:B$} //number of batches 
        \State $\mathcal{B}$ = minibatch sample of $\mathcal{T}_{k}$
        \For{$m=1:M$} 
            \State Sample a noise vector $\epsilon_j\backsim\mathcal{E}$ where $-b\leq\epsilon_j\leq b$
            \State Perturb the feature extractor parameters $\Theta=\{\Phi+\epsilon_j,\Psi\}$
            \For{$x$ in $\mathcal{B} $}
                \State Randomly augment $x$ and $x+$
                \State Compute cluster centers $\mathcal{C}$
                \State Compute loss $\mathcal{L}_{psl}$ and $\mathcal{L}_{psa}$ as per Eq.(2) and (3)
            \EndFor
            \State Compute the base loss $\mathcal{L}_{base}$ as per Eq.(5)
            % \State Compute gradient $\nabla$
            \State Update the network parameters $\{\Phi,\Psi\}$ 
            \State Update the network parameters $\Phi'$ using moving average
            \State Reset the network parameters $\Phi$
            
        \EndFor
    \EndFor
\EndFor

\item[\hspace{5 mm} //Few Shot Learning Phase $k>1$]
\For{$k=2:K$}
    \For{$e=1:E$}
        \State $PseudoLabel= KMeans\_Clustering()$
        \For{$batch=1:B$} // number of batches
            \State $\mathcal{B}$ = minibatch sample of $\mathcal{T}_{k}$
            \For{$x$ in $\mathcal{B} $}
                \State Randomly augment $x$ and $x+$
                \State Compute cluster centers $\mathcal{C}$
                \State Compute loss $\mathcal{L}_{psl}$ and $\mathcal{L}_{psa}$ as per Eq.(2) and (3)
            \EndFor
            \State $\mathcal{B}'=\emptyset$
            \For{$i'=1:S$}
                \State Generate a synthetic sample $\hat{z}_{i'}$ as per Eq.(6)
                \State Apply the transformation module $\hat{\hat{z}}=\kappa_{W_T}(\hat{z}_{i'})$  
                \item [\hspace{24.0 mm} to induce a unbiased synthetic sample $\hat{\hat{z}}_{i'}$]
                \State $\mathcal{B}'=\mathcal{B}'\cup \{\hat{\hat{z}}_{i'},c_i\}$
            \EndFor
            \State Calculate $\mathcal{L}_{ball}$ as per Eq.(7) with  $\mathcal{B}'$
            \State Calculate $\mathcal{L}_{KL}$ with $\mathcal{B}$
            \State Calculate total loss $\mathcal{L}_{k>1}$ as per Eq.(9)
        \EndFor
    \State Update network parameters $\{\Phi,\Psi\},w_X$ with $\mathcal{L}_{k>1}$
    \State Clamp feature extractor parameters $\Phi^*-b\leq\Phi\leq \Phi^*+b$ to meet 
    \item  [\hspace{15 mm} the b-flat-wide local minima conditions]
    \State Update the network parameters $\Phi'$ using moving average
    \EndFor
\EndFor
\end{algorithmic}
\end{algorithm*}

\section{Complexity Analysis}\label{unisa_comp}
Suppose that $\mathcal{T}_k$ is the training data for task $k \in [1,K]$, $N_k$ is the size of $\mathcal{T}_k$, and  $N_{k\mathcal{B}}=|\mathcal{B}|$ is the size of a batch sample of $\mathcal{T}_k$ . Following the UNISA algorithm as presented in algorithm \ref{algorithm1}, UNISA has several atomic operations i.e. generating pseudo label (line 4 and 24) that costs $O(N_k)$, noise sampling and perturbation (line 8-9) that costs $O(1)$, noise line 8) that costs $O(1)$, augmentation (line 11 and 28) that costs $O(1)$, computing cluster center (line 12 and 29) that costs $O(1)$, computing $\mathcal{L}_{psa}$ and $\mathcal{L}_{psl}$ (line 13 and 30) that costs $O(1)$, generating, transforming, and merging synthetic samples (line 34-36) that costs $O(1)$, computing $\mathcal{L}_{ball}$ and $\mathcal{L}_{KL}$ (line 38-39) that costs $O(N_{k\mathcal{B}})$, computing total loss (line 15 and 40) that costs $O(1)$, and finally updating, clamping and resetting network parameters (line 16-18 and 42-44) that costs $O(1)$. Knowing these atomic operations cost, then the complexity of our proposed method can be expressed as:
\begin{equation} \label{}
\small
    O(UNISA) = O(Base Task) + O(Few Shot Task) 
\end{equation}
\begin{equation} \label{}
\small
\begin{split}
    O(Base Task) = & E.( O(N_1) + B.M.(N_{1\mathcal{B}}.O(1) \\
                   & + N_{1\mathcal{B}}.O(1) + O(1) + O(1)) ) 
\end{split}
\end{equation}
\begin{equation} \label{}
\small
\begin{split}
    O(Base Task) = & E.(O(N_1) + B.M.(N_{1\mathcal{B}}.O(1)) ) 
\end{split}
\end{equation}
\begin{equation} \label{}
\small
\begin{split}
    O(Base Task) = & E.(O(N_1) + O(B.M.N_{1\mathcal{B}}) ) 
\end{split}
\end{equation}
Please note that $B.N_{1\mathcal{B}} = N_1$. Therefore equation above can be derived into 
\begin{equation} \label{}
\begin{split}
    O(Base Task) = & E.(O(N_1) + O(M.N_1))
\end{split}
\end{equation}
\begin{equation} \label{}
\begin{split}
    O(Base Task) = E. O(M.N_1) = O(E.M.N_1)
\end{split}
\end{equation}
Please note that the number of perturbation loops ($M$) is a small integer. Therefore the complexity of the Base task can be derived into:
\begin{equation} \label{}
\begin{split}
    O(Base Task) =  O(E.M.N_1) = O(E.N_1)
\end{split}
\end{equation}

Similarly, the complexity of the few shot tasks can be presented by the following equation:
\begin{equation} \label{}
\small
\begin{split}
    O(Few Shot Task) = &  (K-1).E.(O(N_k) + B.(N_{k\mathcal{B}}.O(1) \\
                   & + S.N_{k\mathcal{B}}.O(1) + O(1) + O(1))) 
\end{split}
\end{equation}
\begin{equation} \label{}
\small
\begin{split}
    O(Few Shot Task) = & (K-1).E.(O(N_k) + B.S.(N_{k\mathcal{B}}.O(1)) ) 
\end{split}
\end{equation}
\begin{equation} \label{}
\small
\begin{split}
    O(Few Shot Task) = & (K-1).E.(O(N_k) + O(B.S.N_{k\mathcal{B}})) 
\end{split}
\end{equation}

Similar to the base task, please note that $B.N_{k\mathcal{B}} = N_k$. Therefore equation above can be derived into 
\begin{equation} \label{}
\small
\begin{split}
    O(Few Shot Task) = & (K-1).E.(O(N_k) + O(S.N_{k})) 
\end{split}
\end{equation}
\begin{equation} \label{}
\small
\begin{split}
    O(Few Shot Task) = & (K-1).E.O(S.N_{k})
\end{split}
\end{equation}
\begin{equation} \label{}
\small
\begin{split}
    O(Few Shot Task) = O(E.S.(K-1).N_k) 
\end{split}
\end{equation}
The number of augmented feature ($S$) is a small integer, then the complexity of few shot tasks can be derived into:
\begin{equation} \label{}
 \begin{split}
    O(Few Shot Task) = O(E.S.(K-1).N_k) = O(E.(K-1).N_k) 
\end{split}
\end{equation}
Then the complexity of UNISA can be expressed as
\begin{equation} \label{}
\small
    O(UNISA) = O(Base Task) + O(Few Shot Task) 
\end{equation}
\begin{equation} \label{}
\small
    O(UNISA) =  O(E.N_1) +  O(E.(K-1).N_k) 
\end{equation}
\begin{equation} \label{}
\small
    O(UNISA) =  O(E.(N_1+(K-1).N_k)) 
\end{equation}
Please note that $N_1+(K-1).N_k=N$, where $N$ is the size of the accumulated dataset across all tasks. Then the complexity of UNISA can be derived into:
\begin{equation} \label{}
\small
    O(UNISA) =  O(E.N) 
\end{equation}
Following the standard training of deep learning where number of epoch $E$ is a constant, then the complexity of UNISA can be simplified as:
\begin{equation} \label{}
\small
    O(UNISA) =  O(N) 
\end{equation}
where $N$ is the size of the accumulated dataset across all tasks.

% \fi
\IEEEpeerreviewmaketitle

% \bibliographystyle{IEEEtran}
% \bibliography{references}



%\vfill

